%% file: Constrained_Multi-shape_Evolution_for_Overlapping_Cytoplasm_Segmentation.tex
\documentclass[journal]{IEEEtran}

\usepackage{graphicx}
\usepackage{subfigure}
\usepackage{amsmath}
\usepackage{amssymb}
\usepackage{bbm}
\usepackage{booktabs}
\usepackage{tabularx}
\DeclareMathOperator*{\argmin}{argmin}
\DeclareMathOperator*{\argmax}{argmax}
\begin{document}

\title{Constrained Multi-shape Evolution for Overlapping Cytoplasm Segmentation}

\author{Youyi~Song,
        Lei~Zhu,
        Baiying~Lei,~\IEEEmembership{Senior Member,~IEEE},
        Bin~Sheng,
        Qi~Dou,~\IEEEmembership{Member,~IEEE},
        Jing~Qin$^*$,~\IEEEmembership{Member,~IEEE},
        Kup-Sze~Choi,~\IEEEmembership{Member,~IEEE}\vspace{-14pt}

\thanks{Y. Song, J. Qin and K. S. Choi are with Center for Smart Health, School of Nursing, The Hong Kong Polytechnic University, Hung Hom, Hong Kong.}

\thanks{L. Zhu and Q. Dou are with Department of Computer Science and Engineering, The Chinese University of Hong Kong, Shatin, Hong Kong.}

\thanks{B. Lei is with School of Biomedical Engineering, Shenzhen University, Shenzhen, China.}

\thanks{B. Shen is with School of Mechanical Engineering, Shanghai Jiao Tong University, Shanghai, China.}

\thanks{J. Qin is the corresponding author (e-mail: harry.qin@polyu.edu.hk).}}

\markboth{}{}

\maketitle

\input{Abstract}

\IEEEpeerreviewmaketitle

\input{Introduction}

\input{RelatedWork}

\input{ConstrainedMultishapeEvolution}

\input{Experiments}

\input{Discussion}

\input{Conclusion}

\input{Appendix}

\input{Bibliography}
\end{document}

%% file: Abstract.tex
\begin{abstract}
  Segmenting overlapping cytoplasm of cells in cervical smear images is a clinically essential task, for quantitatively measuring cell-level features in order to diagnose cervical cancer.
  This task, however, remains rather challenging, mainly due to the deficiency of intensity (or color) information in the overlapping region.
  Although shape prior-based models that compensate intensity deficiency by introducing prior shape information (shape priors) about cytoplasm are firmly established, they often yield visually implausible results, mainly because they model shape priors only by limited shape hypotheses about cytoplasm, exploit cytoplasm-level shape priors alone, and impose no shape constraint on the resulting shape of the cytoplasm.
  In this paper, we present a novel and effective shape prior-based approach, called constrained multi-shape evolution, that segments all overlapping cytoplasms in the clump simultaneously by jointly evolving each cytoplasm's shape guided by the modeled shape priors.
  We model local shape priors (cytoplasm--level) by an infinitely large shape hypothesis set which contains all possible shapes of the cytoplasm.
  In the shape evolution, we compensate intensity deficiency for the segmentation by introducing not only the modeled local shape priors but also global shape priors (clump--level) modeled by considering mutual shape constraints of cytoplasms in the clump.
  We also constrain the resulting shape in each evolution to be in the built shape hypothesis set, for further reducing implausible segmentation results.
  We evaluated the proposed method in two typical cervical smear datasets, and the extensive experimental results show that the proposed method is effective to segment overlapping cytoplasm, consistently outperforming the state-of-the-art methods.
\end{abstract}


%% file: Introduction.tex
\section{Introduction}

\IEEEPARstart{T}{he} high-quality screening of cervical cancer has significantly reduced the incidence and mortality of cervical cancer\cite{saslow2012american}---the fourth most common cause of cancer and the fourth most common cause of death from cancer in women all over the world \cite{WHO}.
Cervical cancer screening, more specifically, aims at evaluating the presence or absence of cervical cancer, under a microscope via examining the abnormality extent of each cervical cell that is sampled from the cervix and has been deposited onto a glass slide.

\begin{figure}[!t]
  \centering
  \setlength{\abovecaptionskip}{0pt}
  \setlength{\belowcaptionskip}{0pt}
  \subfigure[]{\includegraphics [width = 1.1in, height = 1.35in]{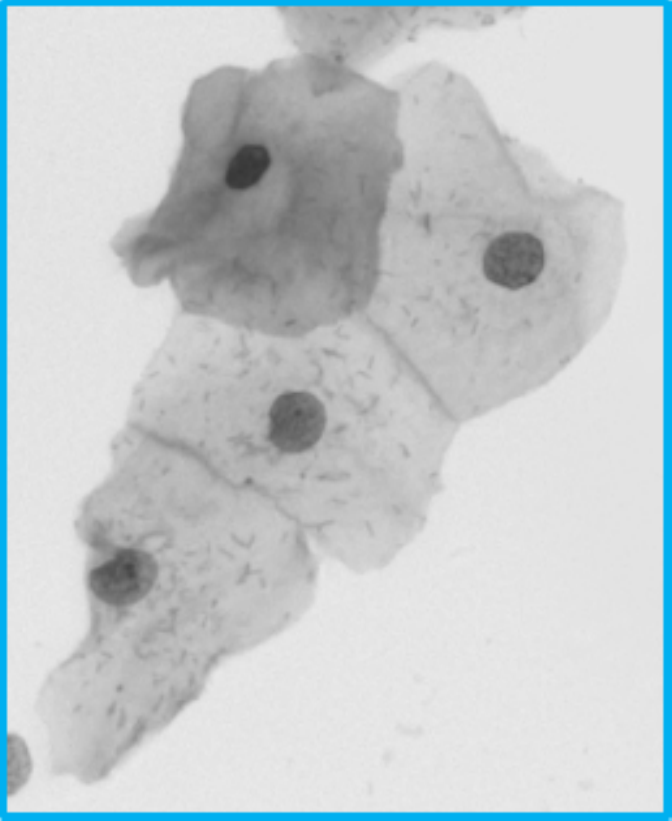}}
  \subfigure[]{\includegraphics [width = 1.1in, height = 1.35in]{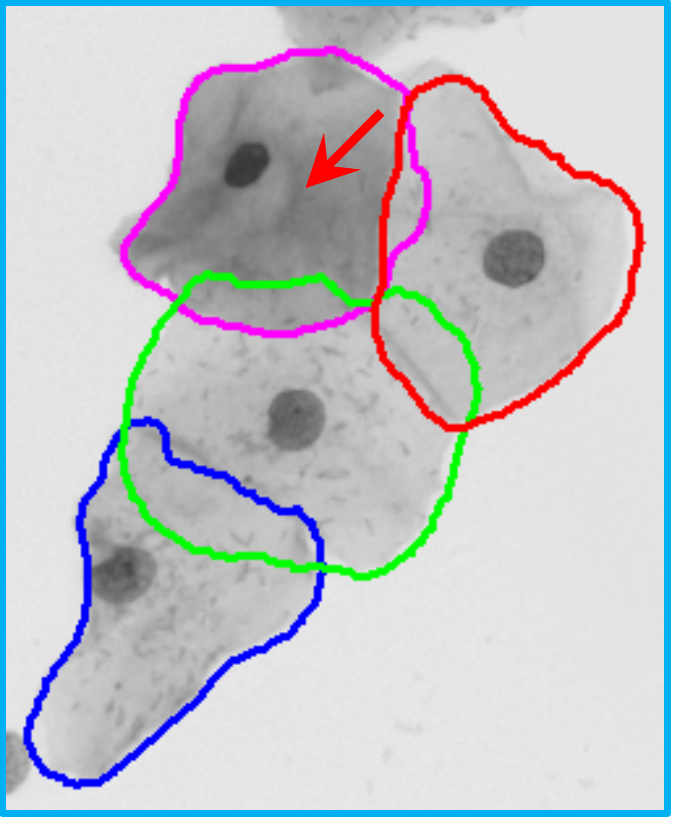}}
  \subfigure[]{\includegraphics [width = 1.1in, height = 1.35in]{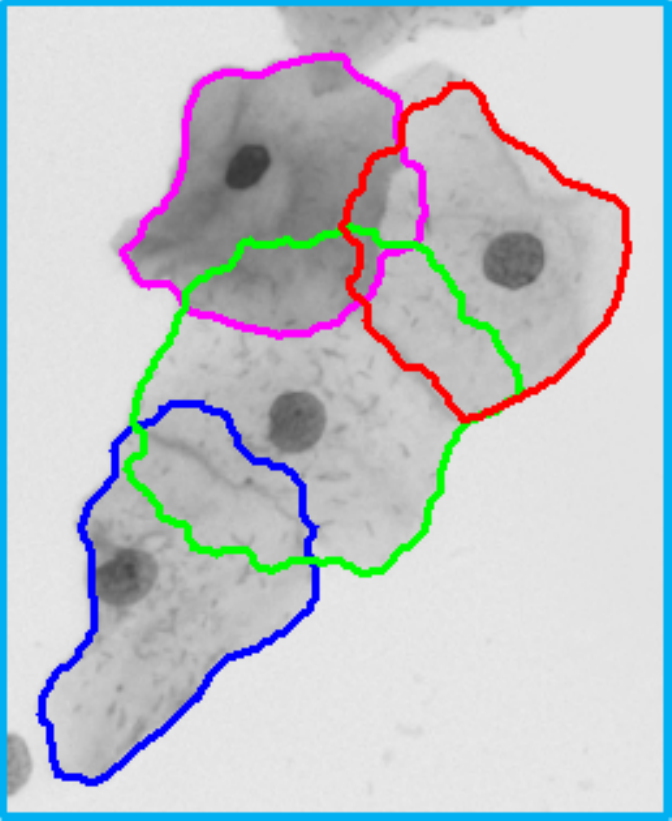}}
  \subfigure[]{\includegraphics [width = 1.1in, height = 1.35in]{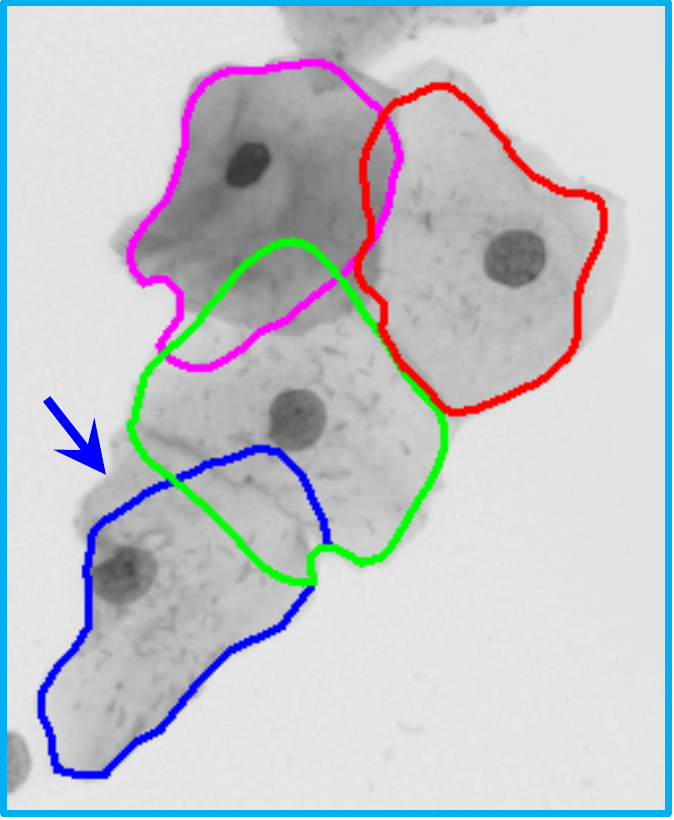}}
  \subfigure[]{\includegraphics [width = 1.1in, height = 1.35in]{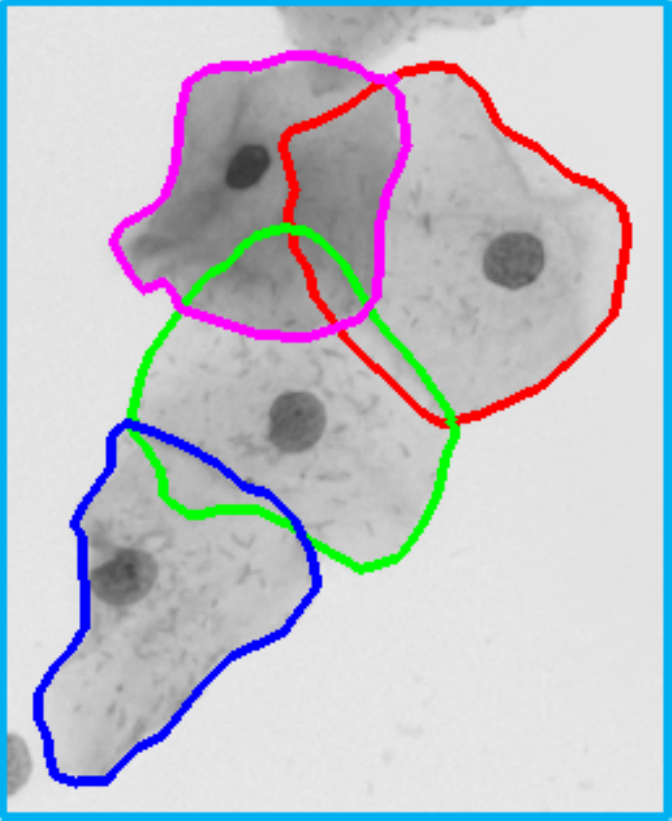}}
  \subfigure[]{\includegraphics [width = 1.1in, height = 1.35in]{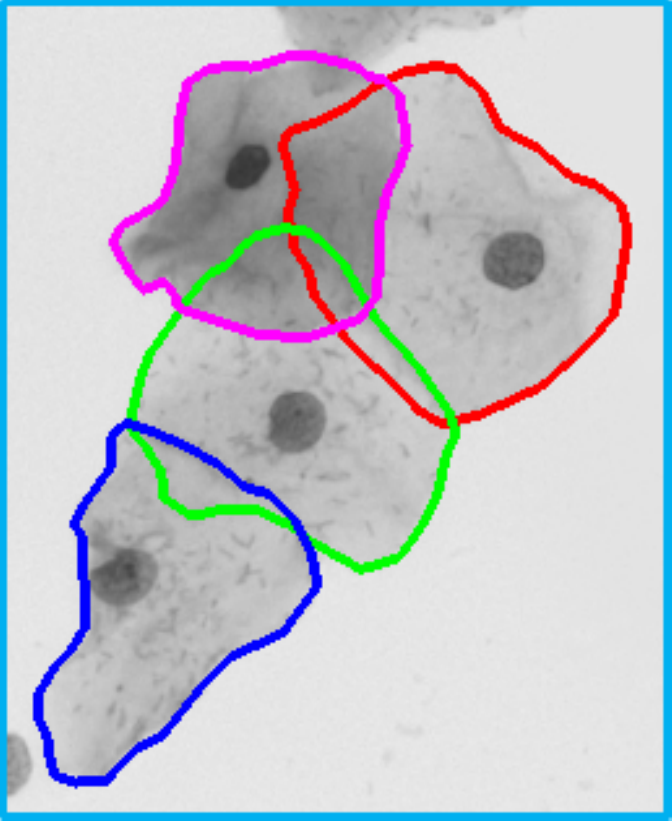}}
  \caption{A visual comparison of different segmentation methods on a typically difficult case: (a) input image, (b)--(e) segmentation results produced by \cite{tareef2018multi}, \cite{lu2015improved}, \cite{song2017accurate}, and the proposed method, respectively, and (f) the ground truth.
  This figure shows that (1) exploiting intensity and spatial information alone \cite{tareef2018multi} cannot address the intensity deficiency issue (see the overlapping region indicated by the red arrow, where the intensity information is deficient for segmenting the cytoplasm depicted by the red curve), and (2) existing shape prior-based approaches \cite{lu2015improved}, \cite{song2017accurate} often yield visually implausible results and their results might be inconsistent with the clump evidence (e.g. segmented boundaries deviate from their desired boundaries in the clump, see such an example indicated by the blue arrow in (d)).}
  \label{Fig.1}
  \vspace{-6pt}
\end{figure}

Segmenting overlapping cytoplasm of cells is a fundamental task for developing automatic cervical cancer screening systems, because in order to examine cell's abnormality extent it is clinically essential to extract cell-level features, e.g., cell's shape, size, and the area ratio of the cytoplasm against the nucleus \cite{kitchener2011automation}.
The clinical evidence has shown that these cell-level features play a crucial role in screening cervical cancer correctly \cite{davey2006effect,guven2014data}, especially in screening early cervical cancer---the only curable stage of cervical cancer at the moment \cite{schiffman2007human}, which is exactly one of the main reasons why so far most research works focus on the segmentation of overlapping cytoplasm for automatically screening cervical cancer rather than directly screening cervical cancer from images by employing a classification model, such as convolutional neural network (CNN).
Segmenting overlapping cytoplasm, on the other hand, is also yet a very challenging task, mainly due to the deficiency of intensity (or color) information in the overlapping region; see Fig.~\ref{Fig.1} for example, where the intensity information in the overlapping region indicated by the red arrow is confusing or even misleading for segmenting the cytoplasm depicted by the red curve.

Early studies \cite{harandi2010automated,plissiti2015segmentation,kumar2016unsupervised,
sulaiman2010overlapping,beliz2011cytology,tareef2018multi,
lee2016segmentation,guan2014accurate,kaur2016curvelet} addressed this segmentation task by exploiting intensity information or together with spatial information among pixels, but these methods tend to fail in overlapping cases where the intensity deficiency issue arises, as shown in Fig.~\ref{Fig.1} (b).
Recently, shape prior-based models \cite{nosrati2015segmentation,nosrati2014variational,lu2015improved,
islam2015multi,song2017accurate,tareef2017automatic,song2016segmenting,
tareef2017optimizing,song2018automated,song2019segmentation} aims at compensating intensity deficiency by introducing prior shape information (shape priors) about cytoplasm into shape-aware segmentation frameworks, say the level set model, for segmenting overlapping cytoplasm.
However, they restrict to modeling shape priors by limited shape hypotheses about cytoplasm, exploiting cytoplasm-level shape priors alone, and imposing no shape constraint on the resulting shape of the cytoplasm.
They hence often cannot provide sufficient prior shape information to pinpoint occluded boundary parts of the overlapping cytoplasm, and also often lead the segmented cytoplasm to an implausible shape, as shown in Fig.~\ref{Fig.1} (c) and (d), the results produced by \cite{lu2015improved} and \cite{song2017accurate} respectively.

In this paper, we present a novel and effective shape prior-based segmentation method, called constrained multi-shape evolution, that segments all overlapping cytoplasms in the clump simultaneously by jointly evolving each cytoplasm's shape guided by the modeled shape priors.
We model local shape priors (cytoplasm-level) by an infinitely large shape hypothesis set which is built by leveraging statistical shape information of cytoplasm.
We also develop a learning algorithm to learn the importance of each shape example in the shape statistics computation, such that the built shape hypothesis set can contain all possible shapes of the cytoplasm.
In the shape evolution, we compensate intensity deficiency for the segmentation by introducing not only the modeled local shape priors but also global shape priors (clump--level) modeled by considering mutual shape constraints of cytoplasms in the clump.
We also constrain the resulting shape in each evolution to be in the built shape hypothesis set, for further reducing implausible segmentation results.

We extensively evaluated the proposed method on two typical cervical smear datasets, and compared it with several state-of-the-art methods.
The experimental results show that the proposed method can effectively cope with the intensity deficiency issue, consistently outperforming the state-of-the-arts; see Fig.~\ref{Fig.1} (e) of our result for example.
We summarize the contributions of our work as follows.
{\begin{itemize}
  \item We propose a novel constrained multi-shape evolution algorithm that is able to effectively model local (cytoplasm-level) and global (clump-level) shape priors for segmenting overlapping cytoplasm.
  \item The proposed constrained multi-shape evolution algorithm can effectively address the intensity deficiency issue (the most difficult problem in the overlapping cytoplasm segmentation task), yielding the state-of-the-art segmentation performance.
\end{itemize}

%% file: RelatedWork.tex
\section{Related Work}

Overlapping cytoplasm segmentation, traditionally, is done by leveraging intensity information or together with the spatial information among pixels in the clump.
This purpose is commonly achieved by extending the classic segmentation models, e.g. thresholding \cite{harandi2010automated,plissiti2015segmentation,kumar2016unsupervised}, watershed \cite{sulaiman2010overlapping,beliz2011cytology,tareef2018multi}, graph-cut \cite{lee2016segmentation}, and morphological filtering \cite{guan2014accurate,kaur2016curvelet}.
They, in principle, assume that intensity information is adequate to identify the occluded boundary parts.
However, this assumption is not always true; in fact, intensity information in the overlapping region is often confusing or even misleading (see Fig.~\ref{Fig.1}).

In order to alleviate the intensity deficiency issue, shape prior-based methods \cite{nosrati2015segmentation,nosrati2014variational,lu2015improved,
islam2015multi,song2017accurate,tareef2017automatic,song2016segmenting,
tareef2017optimizing,song2018automated,song2019segmentation} have shown good segmentation performance by introducing prior shape information (shape priors) to guide the segmentation procedure.
They model shape priors by either a simple shape assumption (e.g. cytoplasm has a star shape \cite{nosrati2015segmentation} or an elliptical shape \cite{nosrati2014variational,lu2015improved,islam2015multi}) or a shape example matched from the pre-collected shape set \cite{song2017accurate,tareef2017automatic,song2016segmenting,
tareef2017optimizing,song2018automated,song2019segmentation}.
They then segment overlapping cytoplasm by still exploiting intensity information but requiring the segmented result to have a shape that is similar with the modeled shape priors as much as possible.
This is often done by employing an active contour model \cite{kass1988snakes} or level set model \cite{chan2001active,li2010distance}, where shape priors are designed as a regularization term in the energy functional whose minimal (or conversely maximal) value is assumed to be attained by the segmentation function that yields the best segmentation result.

Although improving the segmentation accuracy over traditional works, these existing shape prior-based methods \cite{nosrati2015segmentation,nosrati2014variational,lu2015improved,
islam2015multi,song2017accurate,tareef2017automatic,song2016segmenting,
tareef2017optimizing,song2018automated,song2019segmentation} suffered from three main shortcomings.
First, they model shape priors by the limited shape hypotheses, e.g. shape assumptions or collected shape examples, making their modeled shape priors usually insufficiently to pinpoint the occluded boundary parts of the cytoplasm whose shape cannot be well approximated by the specified shape hypothesis.
Second, they evolve cytoplasm's shape by using local shape priors alone (prior shape information about only the individual cytoplasm shape), lack of considering the shape relationship between cytoplasms in the clump, which makes their segmentation results often inconsistent with the clump evidence; e.g. segmented cytoplasms' boundaries deviate from their desired boundaries in the clump (see Fig.~\ref{Fig.1} (d) for example).
Third, in the shape evolution, they do not impose any shape constraint on the resulting shape, though they require the resulting shape to be similar with the modeled shape priors as much as possible.
They, in fact, are to find a suitable compromise between the intensity evidence and the local shape priors by a balance parameter.
As a result, when the intensity evidence contradicts with the local shape priors, they often yield an implausible segmentation result, as shown in Fig.~\ref{Fig.1} (c) and (d).

In this paper, we model local shape priors by leveraging statistical shape information to build a shape space in which shape hypotheses of cytoplasm are infinitely large.
In order to make the built shape hypotheses set (sampled from the built shape space) can contain all possible shapes of cytoplasm, we learn shape example's importance in the shape statistics computation.
In the shape evolution, besides considering local shape priors, we also exploit global shape priors by first jointly evolving cytoplasms' shape together and second explicitly enforcing the consistency between the segmentation results and the clump evidence.
We finally require the resulting shape in each evolution to be in the built shape hypothesis set, for further reducing the implausible segmentation results.

%% file: ConstrainedMultishapeEvolution.tex
\section{Constrained Multi-shape Evolution}

The main difficulty of overlapping cytoplasm segmentation, as we mentioned before, is the deficiency of intensity information in the overlapping region.
As shown in Fig.~\ref{Fig.1}, intensity information in the overlapping region can be confusing or even misleading.
On the other hand, the whole clump boundary is relatively clear in the sense of the intensity contrast (see Fig.~\ref{Fig.1}).
Intuitively, one may wonder if it is possible to segment overlapping cytoplasms by exploiting the clear clump boundary information while ignoring intensity information in the overlapping region.

We will show such a solution is possible, if we have a sufficiently large shape hypothesis set such that each cytoplasm's shape can be well approximated by some shape hypothesis in the set, because under this condition we can cast the segmentation problem into a shape hypotheses selecting problem, by which the clump boundary information can be exploited more easily and efficiently.
The proposed segmentation method accordingly has two core components.
One is to build the shape hypothesis set, with the goal of making all possible shapes of cytoplasm to be contained in the built shape hypothesis set.
Another is to select shape hypotheses, aiming at selecting an optimal shape hypothesis for each cytoplasm for the segmentation, according to the clump boundary information.
For the sake of the presentation, in what follows, we first present the shape hypotheses selecting component, and then present the shape hypothesis set building component, since in order to present the latter component clearly we need to know something about the former component.

\subsection{Multi-shape Hypotheses Selecting}

\begin{figure}[!b]
  \centering
  \setlength{\abovecaptionskip}{0pt}
  \setlength{\belowcaptionskip}{0pt}
  \subfigure[]{\includegraphics [width = 1.6in, height = 1in]{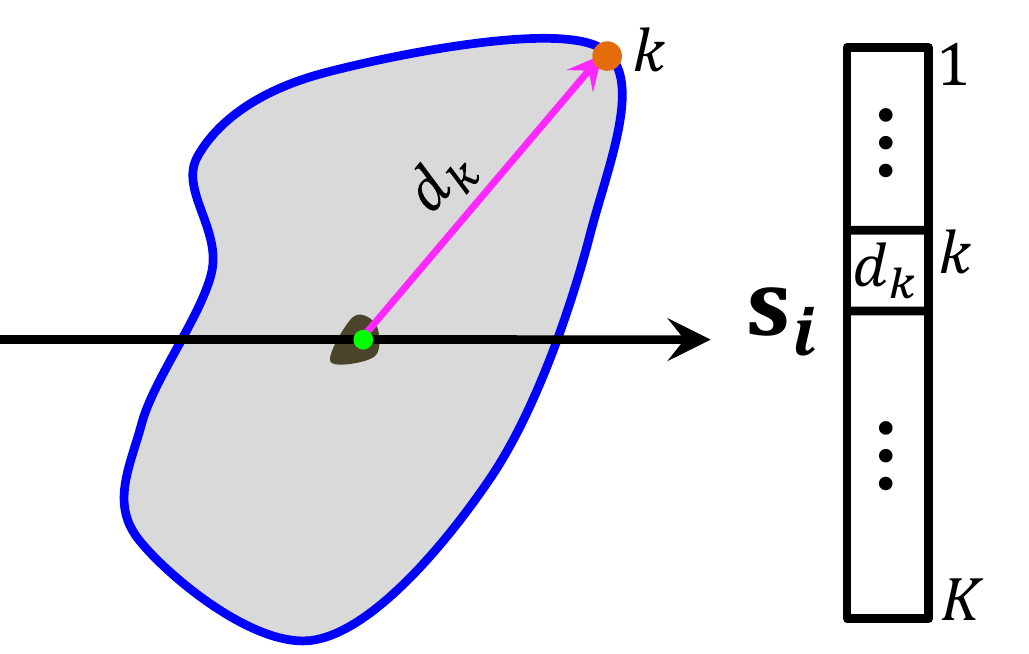}}
  \subfigure[]{\includegraphics [width = 1.6in, height = 1in]{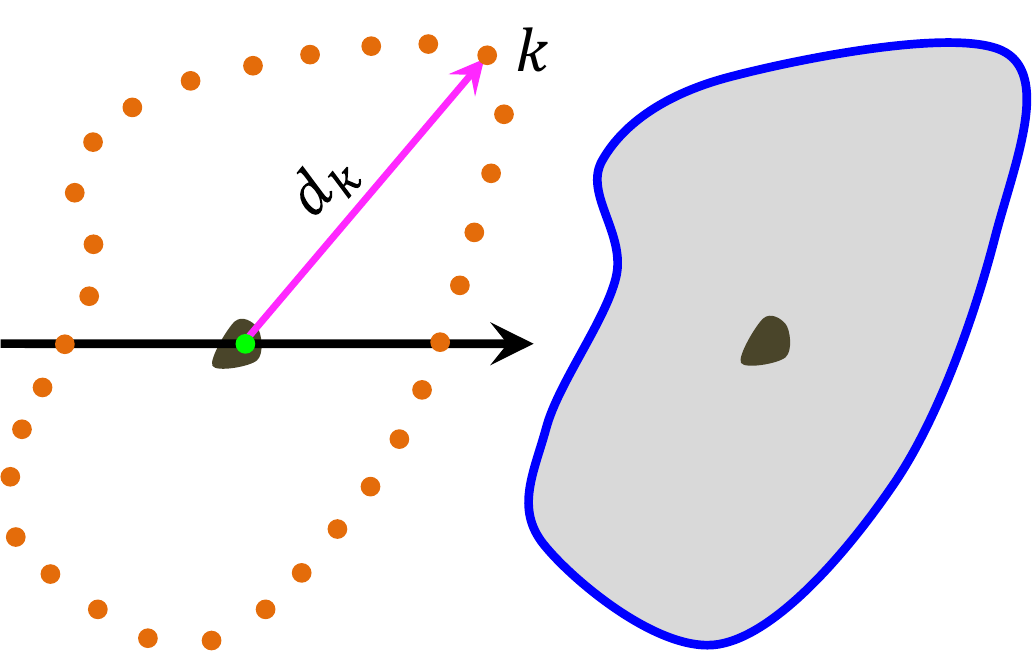}}
  \caption{Graphical depiction of how to use a vector to represent cytoplasm's shape (a) and how to use this vector to produce the segmentation result (b). (a) for each cytoplasm $i$, we sample $K$ boundary points with the regular angle interval to construct its shape vector $\mathrm{\textbf{s}}_i$; more specifically, for the $k$th boundary point, we store its distance $d_k$ to the nucleus' centroid at the $k$th entry of $\mathrm{\textbf{s}}_i$. (b) Given a shape hypothesis $\mathrm{\textbf{s}}_i$, we first locate the corresponding $K$ boundary points in the image, and then fill in the region outlined by these $K$ points as the segmentation result.}
  \label{Fig.2}
\end{figure}

At the moment, we assume that the shape hypothesis set $\boldsymbol{\mathcal{H}}=\{\mathrm{\textbf{s}}_i:\mathrm{\textbf{s}}_i=\boldsymbol{\mu}+\mathrm{\textbf{M}}\mathrm{\textbf{x}}_i, i\in\mathbb{N}\}$ has been built.
Each shape hypothesis $\mathrm{\textbf{s}}_i\in\mathbb{R}^{K\times 1}$ is parameterized by $\textbf{x}_i\in\mathbb{R}^{t\times 1}$, and stores $K$ boundary points' information of the cytoplasm to represent this cytoplasm's shape.
As shown in Fig.~\ref{Fig.2} (a), the $k$th entry of $\textbf{s}_i$ stores the distance value ($d_k$) of the $k$th boundary point; that is $\textbf{s}_i(k)=d_k$.
We defer to discuss $\boldsymbol{\mu}\in\mathbb{R}^{K\times 1}$ and $\mathrm{\textbf{M}}\in\mathbb{R}^{K\times{t}}$ until in the next subsection.
In this subsection, we focus on how to select an appropriate shape hypothesis $\textbf{s}_i$ from $\boldsymbol{\mathcal{H}}$ by finding an optimal value of $\textbf{x}_i$ for segmenting the overlapping cytoplasm $i$ in an input image.

\begin{figure*}[!t]
  \centering
  \setlength{\abovecaptionskip}{-2pt}
  \setlength{\belowcaptionskip}{-6pt}
  \subfigure[]{\includegraphics [width = 1.6in, height = 1.96in]{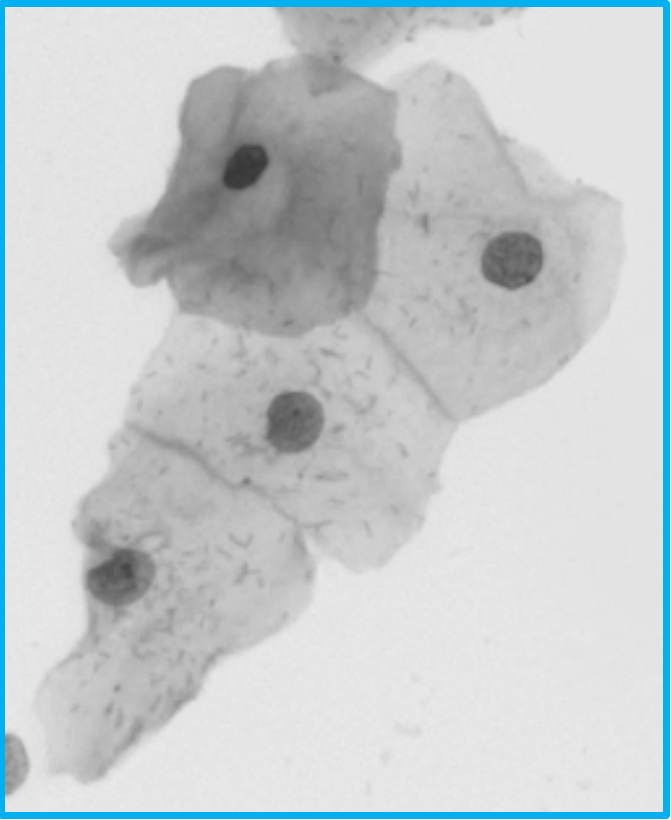}}
  \subfigure[]{\includegraphics [width = 1.6in, height = 1.96in]{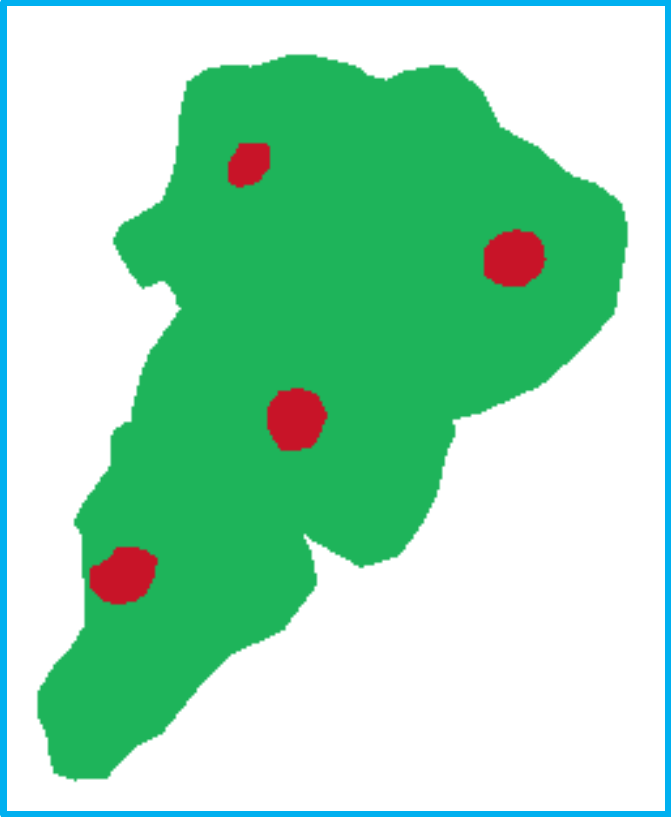}}
  \subfigure[]{\includegraphics [width = 2in, height = 1.96in]{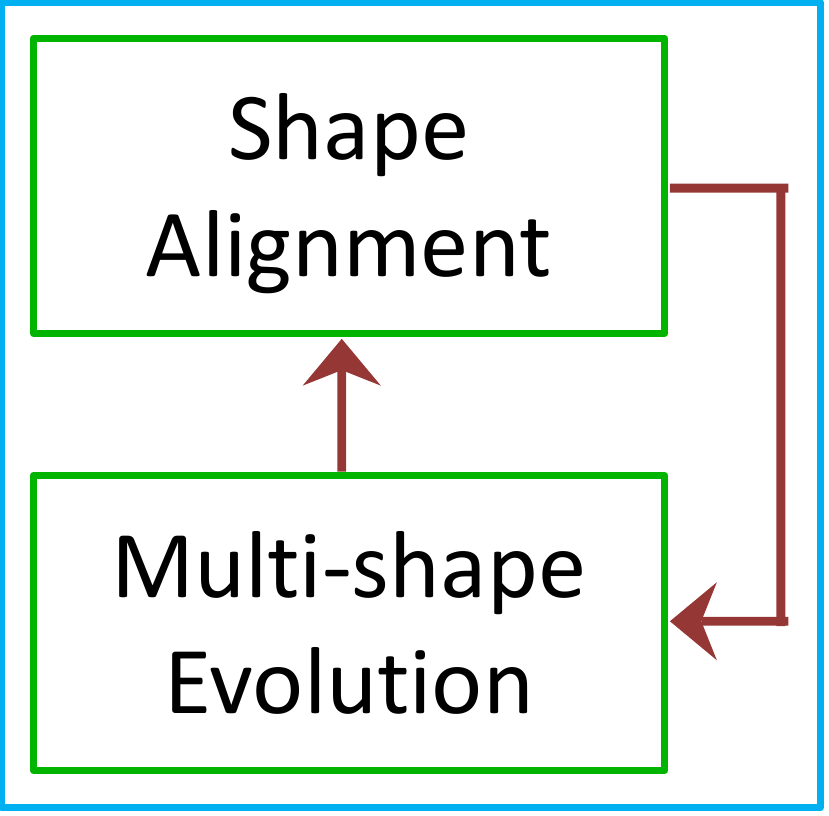}}
  \subfigure[]{\includegraphics [width = 1.6in, height = 1.96in]{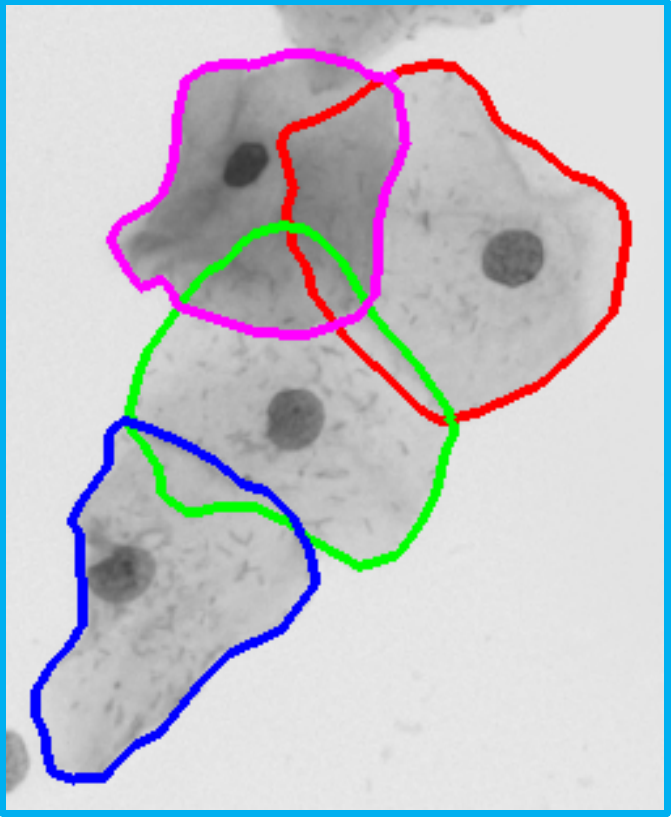}}
  \caption{The pipeline of our multi-shape hypotheses selecting algorithm. Given an input image (a), we first segment the clump area (b), then jointly select the most appropriate shape hypothesis for all cytoplasms in an iterative manner (c), and finally produce the segmentation result according to the selected hypotheses (e).}
  \label{Fig.3}
\end{figure*}

Fig.~\ref{Fig.3} illustrates the pipeline of our multi-shape hypotheses selecting algorithm.
It includes three steps: (1) clump area segmentation, segmenting the clump and the nuclei (the green and red areas in Fig.~\ref{Fig.3} (b)), for providing the clump evidence, (2) shape alignment, producing the segmentation result of cytoplasm $i$ according to $\textbf{s}_i$, and (3) multi-shape evolution, selecting the most appropriate shape hypothesis for all cytoplasms in the clump jointly by progressively finding the optimal value of $\textbf{x}_i$ in an iterative manner.
In what follows, we will present each step in great detail.

\vspace{10pt}\noindent\textbf{Clump Area Segmentation:} We employ a multi-scale convolutional neural network (CNN) reported in \cite{song2015accurate} to segment cytoplasm and nuclei areas (see Fig.~\ref{Fig.3} (b)).
This network classifies each pixel in the image into three classes: nucleus, cytoplasm, or background.
It contains three parallel CNNs, each of which has a different scale in the sense of the resolution of the input patch.
It then fuses three CNNs' outputs for the classification, which helps to capture more contextual information from different scales.
It finally employs the Markov Random Field technique to further refine the segmentation results.
Note that all implementations are the same, so for more technical details please refer to \cite{song2015accurate}.

\vspace{10pt}\noindent\textbf{Shape Alignment:} Given the segmented clump area, and the shape hypothesis $\mathrm{\textbf{s}}_i$, our goal here is to produce the segmentation result of cytoplasm $i$; as mentioned before, how to get $\mathrm{\textbf{s}}_i$ is the goal of the multi-shape evolution step.

We start by locating cytoplasm $i$ by using the location information of nuclei in the segmented clump area.
Since each cervical cell only has a nucleus and a cytoplasm and the nucleus is surrounded by the cytoplasm (see Fig.~\ref{Fig.3} (a)), we use the centroid of nucleus $i$ as the centroid of cytoplasm $i$.
Here we assume that nuclei are not overlapping.
We observed that this assumption is true in the most cases (see Fig.~\ref{Fig.3} (a) for example), and we will discuss how to do when this assumption is not true in the Discussion section (Sec. V).

We now consider how to use $K$ boundary points' information in $\mathrm{\textbf{s}}_i$ to produce the segmentation map of cytoplasm $i$.
For each cytoplasm $i$, we first generate a segmentation map that has the same size with the input image and each pixel in it has a value of $0$.
Next, we locate the corresponding $K$ boundary points in the generated segmentation map by checking the distance of pixels to the cytoplasm's centroid, and fill in the region outlined by these points as the segmentation result of cytoplasm $i$, as shown in Fig.~\ref{Fig.2} (b); pixels in the filled region has a value of $1$.

We finally adjust the scale and rotation of the foreground region, for further prompting the consistence of the segmentation result and the clump evidence.
Note that although the scale and rotation theoretically can be adjusted by selecting a more appropriate $\textbf{s}_i$, it is relatively difficult to implement and by doing so the selecting performance in practice might not be so satisfactory as expected.
We here adopt a simple way that finds the scale ($r_i$) and rotation ($\theta_i$) coefficients by
\begin{equation}
  \begin{aligned}
       \argmax(B_i\cap{B_c}), \text{~s.t.~} B_i\subset{B_c},
  \end{aligned}
\label{Eq.alignment}
\end{equation}
where $B_i$ stands for the adjusted foreground region, rotating the original foreground region with the angle of $\theta_i$ and scaling with the figure of $r_i$.
$B_c$ means the segmented clump area; note that here we re-labeled the cytoplasm and nucleus pixels as $1$, only the background pixels as $0$.

As we can see in Eq.~\ref{Eq.alignment}, we require that the adjusted region $B_i$ has to be inside $B_c$.
Without this requirement, we observed from experiments that the adjusted region is encouraged to be consistent with the whole clump area rather than the cytoplasm itself, deviating from our purpose.
Note that we solve Eq.~\ref{Eq.alignment} by employing the grid search algorithm \cite{lavalle2004relationship}.
We below shall present how to select the shape hypothesis $\textbf{s}_i$.

\vspace{10pt}\noindent\textbf{Multi-shape Evolution:} Our goal here is to select the most appropriate shape hypothesis $\mathrm{\textbf{s}}_i$ for cytoplasm $i$.
For each clump, we select all cytoplasms' shape hypothesis jointly, by which we can exploit more information, prompting the segmentation performance.
As we mentioned before, $\mathrm{\textbf{s}}_i$ is parameterized by $\mathrm{\textbf{x}}_i$, so our goal actually is to find the optimal value of $\mathrm{\textbf{x}}_i$.
Our algorithm starts by setting all $\mathrm{\textbf{x}}_i$ to $\mathrm{\textbf{0}}$, and then finds the optimal value of $\mathrm{\textbf{x}}_i$ in an iterative manner.
In other words, what we actually do is to find a better $\mathrm{\textbf{x}}_i$ at each iteration.

We start by introducing the quantitative measure of `better'.
Since we jointly find the better $\mathrm{\textbf{x}}_i$ for all cytoplasms in the clump, we collectively denote $\{\mathrm{\textbf{x}}_i\}_{i=1}^{N}$ by $\mathrm{\textbf{x}}$, and use the symbol $\mathrm{\textbf{x}}^{k}$ to denote $\mathrm{\textbf{x}}$ at the $k$th iteration.
Here $N$ means the number of cytoplasms in the clump, determined by the nuclei number in the segmented clump area.
Our measure is based on the following function
\begin{equation}
       \mathbb{E}(B_c, \mathrm{\textbf{x}}^k )=\sum_{(x,y)\in\Omega_B}\bigl(B_u(x,y)-B_c(x,y)\bigr)^2,
\label{Eq.discrepancy}
\end{equation}
where $B_c$ is the re-labeled segmented clump area (as mentioned before, all cytoplasm and nucleus pixels have the same value of $1$), $B_u=\bigcup_{i=1}^{N}B_i$ is the union of $N$ cytoplasms' segmentation results produced by the current shape hypotheses $\{\textbf{s}_i\}_{i=1}^{N}$, and $(x, y)$ indicates pixel's coordinate in the image.
Having function $\mathbb{E}$, we say that $\mathrm{\textbf{x}}^{k+1}$ is better than $\mathrm{\textbf{x}}^{k}$ if we have $\mathbb{E}(B_c, \mathrm{\textbf{x}}^{k+1})<\mathbb{E}(B_c, \mathrm{\textbf{x}}^{k})$.

We can see from Eq.~\ref{Eq.discrepancy} that we just check the pixel-wise difference between $B_u$ and $B_c$, without checking the segmentation quality of $\{B_i\}_{i=1}^{N}$.
We also can see that the best $\mathrm{\textbf{x}}^{\star}$ we can obtain is when $\mathbb{E}(B_c, \mathrm{\textbf{x}}^{\star})=0$, and in this ideal case we have $B_u=B_c$.
However, there may exist too many such $\mathrm{\textbf{x}}^{\star}$, and the segmentation results $\{B_i\}_{i=1}^{N}$ produced by some of them may heavily deviate from the real segmentation.
In other words, we have not a sufficiently tight guarantee on the segmentation performance in the overlapping region.

In fact, it is the price that we cannot exploit intensity information in overlapping region.
But it can be addressed by well designing the shape hypothesis set $\boldsymbol{\mathcal{H}}$ which will be discussed in great detail in the next subsection.
At the moment, we focus on how to get a better $\mathrm{\textbf{x}}^{k+1}$ by using information at $k$th iteration.
The mathematical tool we used here is Taylor's Theorem \cite {graves1927riemann} that states that for any $\mathrm{\textbf{p}}$ we have
\begin{equation}
       \mathbb{E}(\mathrm{\textbf{x}}^{k}+\mathrm{\textbf{p}})=
       \mathbb{E}(\mathrm{\textbf{x}}^{k})+
       \triangledown\mathbb{E}(\mathrm{\textbf{x}}^{k})^T
       \mathrm{\textbf{p}}+\frac{1}{2}\mathrm{\textbf{p}}^T
       \triangledown^2\mathbb{E}(\mathrm{\textbf{x}}^{k}+
       \gamma{\mathrm{\textbf{p}}})\mathrm{\textbf{p}},
\label{Eq.optimalBase}
\end{equation}
where $\triangledown$ and $\triangledown^2$ are the gradient and the Hessian, and $\gamma$ is some scalar in the interval $(0,1)$.
It means that we can approximate $\mathbb{E}$ near $\mathrm{\textbf{x}}^{k}$ using information about only the function value, and the first-order and second-order derivatives at $\mathrm{\textbf{x}}^{k}$.
It then allows us to obtain the global minimizer of $\mathbb{E}$ in the region centred at $\mathrm{\textbf{x}}^{k}$ with a radius as $\|\mathrm{\textbf{p}}\|_2$.
Theoretically, the global minimizer is the best $\mathrm{\textbf{x}}^{k+1}$ we can access at the $k$th evolution.

But since we do not know the value of $\gamma$, $\mathbb{E}$ cannot be analytical known.
We hence turn to approximating $\mathbb{E}$ at $\mathrm{\textbf{x}}^{k}$, denoted by $m_k$, as the following way,
\begin{equation}
       m_k(\mathrm{\textbf{p}})=\mathbb{E}(\mathrm{\textbf{x}}^{k})
       +\triangledown\mathbb{E}(\mathrm{\textbf{x}}^{k})^T
       \mathrm{\textbf{p}}+\frac{1}{2}\mathrm{\textbf{p}}^T
       \triangledown^2\mathbb{E}(\mathrm{\textbf{x}}^{k})
       \mathrm{\textbf{p}}.
\label{Eq.modellingFunction}
\end{equation}
It is especially accurate when $\|\mathrm{\textbf{p}}\|_2$ is small; the approximation error is $O(\|\mathrm{\textbf{p}}\|_2^3)$.
We then approximate the minimizer of $\mathbb{E}$ by the minimizer of $m_k$, that is,
\begin{equation}
       \mathrm{\textbf{p}}^*=\argmin_{\mathrm{\textbf{p}}\in\Omega_\mathrm{\textbf{p}}}m_k(\mathrm{\textbf{p}}), \text{~s.t.~} \|\mathrm{\textbf{p}}\|_2\le\Delta_k,
\label{Eq.optimalFunction}
\end{equation}
where $\Delta_k>0$ is the radius that specifies the region in which we search the optimal $\mathrm{\textbf{p}}^*$.
We solve the above equation by employing the trust-region algorithm \cite{nocedal2006nonlinear} that can automatically choose the value of $\Delta_k$ at each iteration.
Once we get $\mathrm{\textbf{p}}^*$, we set $\mathrm{\textbf{x}}^{k+1}$ to  $\mathrm{\textbf{x}}^{k}+\mathrm{\textbf{p}}^*$.

We find the optimal value of $\mathrm{\textbf{x}}$ by repeatedly implementing the above iterative process.
We halt to iterate when the value of $\mathbb{E}$ cannot be decreased or reaches the prescribed threshold $\mathbb{E}^*$, and use the terminated $\mathrm{\textbf{x}}$ to produce the segmentation result, as shown in Fig.~\ref{Fig.3} (e).
Up to now, we have presented all technical details about how we implement the multi-shape hypotheses selecting component for segmenting overlapping cervical cytoplasms.
Below we shall present details about how to build the shape hypothesis set $\boldsymbol{\mathcal{H}}$.

\subsection{Shape Hypothesis Set Building}

As we mentioned before, our shape hypothesis set is built as $\boldsymbol{\mathcal{H}}=\{\mathrm{\textbf{s}}_i:\mathrm{\textbf{s}}_i=\boldsymbol{\mu}+\mathrm{\textbf{M}}\mathrm{\textbf{x}}_i, i\in\mathbb{N}\}$.
Our idea is to leverage cytoplasm's statistical shape information, as suggested in \cite{cootes1995active}.
To do so, we first collect a shape example set; details are given in the Experiments section (Sec. IV).
Each example represents its cytoplasm shape by using the same vector format, as shown in Fig.~\ref{Fig.2}.
The statistical shape information we used here actually is the mean shape $\boldsymbol{\mu}\in\mathbb{R}^{K\times 1}$ of the collected shape examples and the eigenvectors of the collected examples' covariance matrix.
Note that each column of the matrix $\mathrm{\textbf{M}}\in\mathbb{R}^{K\times{t}}$ is a eigenvector; more specifically, $\mathrm{\textbf{M}}=(\mathrm{\textbf{e}}_1\ \mathrm{\textbf{e}}_2\ \cdots\ \mathrm{\textbf{e}}_t)$, where the corresponding eigenvalue $\lambda_1\geq\lambda_2\geq\cdots\geq\lambda_t$.
For more implementation details about how to compute $\boldsymbol{\mu}$ and $\mathrm{\textbf{M}}$, please refer to \cite{cootes1995active}.

Building shape hypothesis set by this way, we can see that each shape hypothesis $\mathrm{\textbf{s}}_i$ in fact is the sum of $\boldsymbol{\mu}$ and a linear combination of first $t$ eigenvectors.
This way is widely used, because by choosing a different combination $\mathrm{\textbf{x}}_i$ we are able to get a different shape hypothesis $\mathrm{\textbf{s}}_i$, and hence we can readily build an infinitely large shape hypothesis set by choosing infinitely many values of $\mathrm{\textbf{x}}_i$.

Its main shortcoming is that the representation capability of the built shape hypothesis set depends on how good of the shape examples we have collected \cite{heimann2009statistical}.
In practice, we often have to manually select the representative shape examples, in a trial and error manner, which is labour-intensive and becoming infeasible when we have to collect infinitely many examples for approximating complex shapes \cite{gurcan2009histopathological}.
In other words, it is difficult or even impractical to collect a set of shape examples in advance by which any unseen cytoplasm's shape can be well approximated by some built shape hypothesis.

One may argue that by the above way it is still possible to approximate any unseen shape by carefully choosing the value of $\mathrm{\textbf{x}}_i$.
However, it is rather difficult to design such a choosing algorithm in the sense of the theoretical difficulty and$\setminus$or computational difficulty.
In order to alleviate the burden on the manual collecting of shape examples, we propose a learning algorithm that aims at learning example's importance in the shape statistics computation, by which the representation capability of the built shape hypothesis set is not so sensitive to which examples we collected.

\vspace{10pt}\noindent\textbf{Learning Shape Example's Importance:} We denote the collected set of $K$ input-output pairs by $\mathcal{D}=\{(B_c^j, \{\mathrm{\textbf{s}}_i^j\}_{i=1}^{N_j})\}_{j=1}^K$, where $B_c^j$ is the clump area segmentation result of the training image indexed by $j$, and $\mathrm{\textbf{s}}_i^j$ stands for the shape vector of cytoplasm $i$ in image $j$ in which $N_j$ cytoplasms appear.
Note that although we use the same symbol $\mathrm{\textbf{s}}_i$ to denote both the shape hypothesis $i$ and the shape example $i$ because they are the same in essence, it is not necessary to upset by this symbol abuse; we will distinguish them when the context is not clear enough.

Our learning algorithm starts by taking out all shape examples $\{\mathrm{\textbf{s}}_i^j\}$ from $\mathcal{D}$.
Each example $\mathrm{\textbf{s}}_i^j$ initially has its importance, denoted by $w_i^j$, as $1$, and we compute the mean shape as
\begin{equation}
       \boldsymbol{\mu}=\frac{1}{W}
       \sum_{j=1}^{K}\sum_{i=1}^{N_j}
       {w_i^j\mathrm{\textbf{s}}_i^j},
\label{Eq.imageU}
\end{equation}
where $W$ is the sum of all $w_i^j$.
We then compute the covariance matrix as the following way
\begin{equation}
       \mathrm{\textbf{M}}_c=\frac{1}{N}\sum_{j=1}^{K}\sum_{i=1}^{N_j}
       (\mathrm{\textbf{s}}_i^j-\boldsymbol{\mu})
       (\mathrm{\textbf{s}}_i^j-\boldsymbol{\mu})^T,
\label{Eq.imageM}
\end{equation}
where $N$ stands for the number of all cytoplasms in $\mathcal{D}$.
The matrix $\mathrm{\textbf{M}}$, as mentioned before, is comprised of the first $t$ eigenvectors of $\mathrm{\textbf{M}}_c$, in a column order with the decreasing eigenvalue.

We then learn the importance $\{w_i^j\}$ as follows.
We first take an input-output pair $(B_c^j, \{\mathrm{\textbf{s}}_i^j\}_{i=1}^{N_j})$ from $\mathcal{D}$.
We then run our multi-shape hypotheses selecting algorithm using current $\boldsymbol{\mu}$ and $\mathrm{\textbf{M}}$, and check the terminated value $\mathbb{E}(B_c^j, \mathrm{\textbf{x}}^{*} \vert \boldsymbol{\mu}, \mathrm{\textbf{M}})$, where $\mathrm{\textbf{x}}^{*}$ means the optimal value of $\textbf{x}$ we can get under the condition of current $\boldsymbol{\mu}$ and $\mathrm{\textbf{M}}$.
We next update $\{w_i^j\}_{i=1}^{N_j}$ one by one at each step.
At each step, $w_i^j$ continues to be increased by $\ell$ at each try if the terminated $\mathbb{E}$ is to be decreasing under the new $\boldsymbol{\mu}$ and $\mathrm{\textbf{M}}$ which are both changed under new $w_i^j$.
Once all $\{w_i^j\}_{i=1}^{N_j}$ have been updated, we take a new input-output pair from $\mathcal{D}$ and implement above updating procedure again.
Once all input-output pairs are used, the updating procedure cycles again through all input-output pairs until there are no longer input-output pair whose terminated $\mathbb{E}$ value can be decreased by altering their examples' importance.

It is worth to note that although our learning algorithm updates the importance by considering only one example at a step, which may `mess up' the terminated $\mathbb{E}$ value of other $\{B_c^j\}$ that are not involved in the current updating step, it turns out that the importance learned by our learning algorithm is guaranteed to arrive at the right importance in the end.
The result holds regardless of which input-output pair $(B_c^j, \{\mathrm{\textbf{s}}_i^j\}_{i=1}^{N_j})$ we choose and which shape example $\mathrm{\textbf{s}}_i^j$ in the input-output pair we choose at each importance updating step.
The formal proof of above statements is presented in the `Appendix A', because it is the heaviest in this paper in terms of mathematical abstraction and it can be safely skipped without losing the `plot'.
Below we just give an intuitive explanation why above statements hold.

For symbol simplicity, we collectively denote all $\{w_i^j\}$ by $\textbf{w}$, and denote the right importance and the learned importance at step $t$ by respectively $\textbf{w}^*$ and $\textbf{w}(t)$.
The right importance $\textbf{w}^*$ means by which for all input-output pairs the terminated $\mathbb{E}$ value reaches its minimal value.
The basic idea behind the proof is to check the angle $\theta(t)$ between $\textbf{w}^*$ and $\textbf{w}(t)$.
We start by looking at
\begin{equation}
       \cos\theta(t)=\frac{\textbf{w}^T(t)\textbf{w}^*}
       {\Vert\textbf{w}(t)\Vert\Vert\textbf{w}^*\Vert}\le 1.
\label{Eq.angle}
\end{equation}
It is always true.
However, if above statements did not hold, meaning that $t$ will be infinitely large, then we will arrive at a contradiction: $\cos\theta(t)>1$.
So $t$ is finite, meaning that after finite $t$ steps the learned importance $\textbf{w}(t)$ makes all input-output pairs get its minimal terminated $\mathbb{E}$ value, which implies $\textbf{w}(t)=\textbf{w}^*$\footnote{Strictly speaking, there may exist more than one right importance $\textbf{w}^*$, so $\textbf{w}(t)$ is just equal to some possible $\textbf{w}^*$, not to all $\textbf{w}^*$.}.

We now explain why the contradiction appears.
From a probabilistic point of view, our updating rule of $\textbf{w}$ can be treated as a random walk \cite{spitzer2013principles}, moving towards $\textbf{w}^*$ at some steps while also being away from $\textbf{w}^*$ at other steps.
If this random walk was unbiased, then $\Vert\textbf{w}(t)\Vert$ typically would grow as $\sqrt{t}$, and $\textbf{w}^T(t)\textbf{w}^*$, though will fluctuate, generally will grow as $\sqrt{t}$ by central limit theorem \cite{rosenblatt1956central}.
However, as shown in the `Appendix A', this random walk is biased, and hence $\textbf{w}^T(t)\textbf{w}^*$ grows linearly with $t$, while on the other hand $\textbf{w}(t)$ is biased away from $\textbf{w}(t-1)$, making $\Vert\textbf{w}(t)\Vert$ grow at most as $\sqrt{t}$.
In this case, if $t$ was infinitely large, then $\cos\theta(t)\propto\frac{t}{\sqrt{t}}=\sqrt{t}=+\infty$.

%% file: Experiments.tex
\section{Experiments}

\subsection{Datasets}

We evaluate the proposed segmentation method on two typical cervical smear datasets that are prepared with different staining manners.
The first dataset is the Pap Stain Dataset prepared by Papanicolaou stain, which is obtained from \emph{ISBI} $2015$ Overlapping Cervical Cytology Image Segmentation Challenge \cite{lu2017evaluation}.
This dataset consists of $8$ publicly available images, and each of them on average has $11$ clumps with $3.3$ cytoplasm in a clump.
The second dataset is the H\&E Stain Dataset prepared by Hematoxylin and Eosin (H\&E) stain, which is obtained from Shenzhen Sixth People's Hospital.
This dataset consists of $21$ images, and each of them on average has $7$ clumps with $6.1$ cytoplasm in a clump.
These two datasets are widely used in the overlapping cervical cytoplasm segmentation studies, such as in \cite{tareef2018multi,lu2015improved,song2017accurate,song2018automated}.
It exactly is the main reason why we choose them to evaluate the proposed method; by doing so, we can get a relatively fair comparison with existing methods.

\subsection{Evaluation Metric}

We first employ four widely-used pixel-level metrics: True Positive Rate ($TPR$), False Positive Rate ($FPR$), True Negative Rate ($TNR$), and False Negative Rate ($FNR$).
$TPR$, also known as Recall or Sensitivity, is the fraction of cytoplasm pixels segmented correctly, while $FPR$ is the fraction of non-cytoplasm pixels segmented erroneously.
$TNR$, also known as Specificity or Selectivity, is the fraction of non-cytoplasm pixels segmented correctly, while $FNR$ is the fraction of cytoplasm pixels segmented erroneously.
We further employ a object-level metric: Dice Similarity Coefficient ($DSC$), which is also widely-used especially in the state-of-the-arts \cite{tareef2018multi,lu2015improved,song2017accurate,song2018automated}. %
It is defined as $2(|C_s\cap{C_g}|)/(|C_s| + |C_g|)$, where $C_s$ and $C_g$ denote the segmentation result and the ground truth of the cytoplasm respectively, and the operation $|C|$ counts the number of cytoplasm pixels in the binary image $C$.
Note that all five metrics take its value in the range of [$0$, $1$], and a better segmentation method has a larger value of $DSC$, $TPR$, and $TNR$, while a smaller value of $FPR$ and $FNR$.

\subsection{Implementation Details}

We employ a $5$-fold cross validation scheme to construct the training set $\mathcal{D}$.
In each part of the validation, we have about $45$ clumps with $221$ cytoplasm on average for the training and about $180$ clumps with $887$ cytoplasm on average for the testing.
The number of cytoplasm in each clump is from $2$ to $13$, with the mean of $4.93$ and the standard deviation of $1.81$.
Different from the conventional cross validation, we here use $1/5$ images for the training and $4/5$ images for the testing, because our primary purpose is to demonstrate that any unseen shape during training can be well approximated by some shape hypothesis in the built set.
By employing the conventional way, it is more likely that some of the testing shapes have been seen during training.

\subsection{Parameters Selection}

Our method has three parameters: (1) $K$, the number of boundary points to represent cytoplasm's shape, (2) $t$, the number of eigenvectors to build the shape hypothesis, and (3) $\mathbb{E^*}$, the prescribed threshold to halt the multi-shape evolution procedure.
We determine their values by mainly considering the segmentation accuracy and computational cost.
We set $K$ to $360$, such that each boundary point represents shape information at each integer angle of the shape; it is intuitively natural.
We also observed from experiments that it also can be set to $180$ or $720$ (or even $90$ or $1080$) without no significant performance varying.
Compared with $360$, a larger value produces a slightly more accurate segmentation result while also consuming more computational cost (a smaller value has an opposite trend).
Regarding the eigenvector number $t$, a large $t$ makes the built shape hypothesis $\mathrm{\textbf{s}}_i$ represent more local shape details, but consumes more computational resources in the shape evolution.
As suggested in \cite{cootes1995active}, we determine $t$ according to the requirement: $(\sum_{i=1}^t \lambda_i/\sum \lambda_i) > 0.995$, and hence the corresponding value of $t$ is $20$ in this work.
As for $\mathbb{E^*}$, a small value usually has a slight improvement on the segmentation accuracy, however it at the same time incurs more computational time.
We hence empirically set $\mathbb{E^*}$ to $5\%$ of the number of pixels in the clump.

\begin{table*}[!t]\small
\setlength{\abovecaptionskip}{0pt}
\caption{Quantitative Segmentation Results Measured by the Pixel-level Metrics}
\vspace{-6pt}
\begin{center}
\begin{tabular}{p{0.55in} p{0.6in} p{0.6in} p{0.56in} p{0.6in}
                p{0.08in} p{0.6in} p{0.6in} p{0.6in} p{0.6in}}
\toprule
\ &\multicolumn{4}{c}{\ Pap Stain Dataset} \ &\multicolumn{4}{c}{\ \ \ \ \ \ \ \ \ \ \ \ \ \ \ \ \ \ \ \ \ \ \ H\&E Stain Dataset}\\

\cmidrule{2-5}
\cmidrule{7-10}

\ &{\ \ \ $TPR$}
  &{\ \ \ $TNR$}
  &{\ \ \ $FPR$}
  &{\ \ $FNR$} &{}
  &{\ \ \ $TPR$}
  &{\ \ \ $TNR$}
  &{\ \ \ $FPR$}
  &{\ \ $FNR$}\\

\midrule
{LSF\cite{lu2015improved}}
      &{0.88$\pm$0.08} &{0.98$\pm$0.01} &{0.02$\pm$0.01}
      &{0.37$\pm$0.19} &{}
      &{0.86$\pm$0.10} &{0.97$\pm$0.01} &{0.03$\pm$0.01}
      &{0.39$\pm$0.21}\\

{MCL\cite{song2017accurate}}
      &{0.87$\pm$0.10} &{0.98$\pm$0.01} &{0.02$\pm$0.01}
      &{0.34$\pm$0.21} &{}
      &{0.85$\pm$0.09} &{0.98$\pm$0.01} &{0.02$\pm$0.01} &{0.36$\pm$0.20}\\

{MPW\cite{tareef2018multi}}
      &{0.90$\pm$0.07} &{\textbf{0.99}$\pm$0.01} &{\textbf{0.01}$\pm$0.01}
      &{0.31$\pm$0.18} &{}
      &{0.89$\pm$0.07} &{\textbf{0.99}$\pm$0.01} &{\textbf{0.01}$\pm$0.01} &{0.32$\pm$0.16}\\

{CF\cite{song2018automated}}
     &{0.92$\pm$\textbf{0.06}} &{\textbf{0.99}$\pm$0.01} &{\textbf{0.01}$\pm$0.01}
     &{0.30$\pm$0.17} &{}
     &{0.90$\pm$\textbf{0.06}} &{\textbf{0.99}$\pm$0.01} &{\textbf{0.01}$\pm$0.01} &{0.31$\pm$0.19}\\

\midrule
{Ours-F} &{0.90$\pm$0.08} &{0.98$\pm$0.01} &{0.02$\pm$0.01}
         &{0.32$\pm$0.17} &{}
         &{0.88$\pm$0.09} &{0.98$\pm$0.01} &{0.02$\pm$0.01} &{0.32$\pm$0.19}\\

{Ours-M} &{0.91$\pm$0.07} &{0.98$\pm$0.01} &{0.02$\pm$0.01}
         &{0.34$\pm$0.15} &{}
         &{0.90$\pm$\textbf{0.06}} &{0.98$\pm$0.01} &{0.02$\pm$0.01} &{0.35$\pm$0.17}\\

{Ours-L} &{0.90$\pm$0.09} &{\textbf{0.99}$\pm$0.01} &{\textbf{0.01}$\pm$0.01}
         &{0.29$\pm$0.18} &{}
         &{0.89$\pm$0.08} &{\textbf{0.99}$\pm$0.01} &{\textbf{0.01}$\pm$0.01} &{0.31$\pm$0.19}\\

\midrule
{Ours} &{\textbf{0.94}$\pm$\textbf{0.06}}
       &{\textbf{0.99}$\pm$0.01}
       &{\textbf{0.01}$\pm$0.01}
       &{\textbf{0.28}$\pm$\textbf{0.13}} &{}
       &{\textbf{0.93}$\pm$0.07}
       &{\textbf{0.99}$\pm$0.01} &{\textbf{0.01}$\pm$0.01} &{\textbf{0.27}$\pm$\textbf{0.15}}\\
\bottomrule
\end{tabular}
\end{center}
\label{TableResultPixel}
\vspace{-6pt}
\end{table*}

\begin{table*}[!t]\small
\setlength{\abovecaptionskip}{0pt}
\caption{Quantitative Segmentation Results Measured by the Object-level Metric ($DSC$) According to the Overlapping Degree}
\vspace{-6pt}
\begin{center}
\begin{tabular}{p{0.55in} p{0.45in} p{0.45in} p{0.45in} p{0.45in}
               p{0.45in} p{0.0in} p{0.45in} p{0.45in} p{0.45in} p{0.45in} p{0.45in}}
\toprule
\ &\multicolumn{5}{c}{\ \ \ Pap Stain Dataset} \ &\multicolumn{5}{c}{\ \ \ \ \ \ \ \ \ \ \ \ \ \ \ \ \ \ \ \ H\&E Stain Dataset}\\

\cmidrule{2-6}
\cmidrule{8-12}

\ &{\ \ \ ($0$, $\frac{1}{4}$)}
  &{\ \ \ [$\frac{1}{4}$, $\frac{2}{4}$)}
  &{\ \ \ [$\frac{2}{4}$, $\frac{3}{4}$)}
  &{\ \ \ [$\frac{3}{4}$, $1$)}
  &{\ \ Overall} &{}
  &{\ \ \ ($0$, $\frac{1}{4}$)}
  &{\ \ \ [$\frac{1}{4}$, $\frac{2}{4}$)}
  &{\ \ \ [$\frac{2}{4}$, $\frac{3}{4}$)}
  &{\ \ \ [$\frac{3}{4}$, $1$)}
  &{\ \ Overall}\\

\midrule
{LSF\cite{lu2015improved}}
      &{0.81$\pm$0.07} &{0.77$\pm$0.06} &{0.74$\pm$0.07}
      &{0.71$\pm$0.10} &{0.77$\pm$0.07} &{}
      &{0.80$\pm$0.08} &{0.75$\pm$0.07} &{0.72$\pm$0.08}
      &{0.69$\pm$0.11} &{0.75$\pm$0.08}\\

{MCL\cite{song2017accurate}}
      &{0.80$\pm$0.05} &{0.79$\pm$0.06} &{0.77$\pm$0.07}
      &{0.72$\pm$0.08} &{0.78$\pm$0.06} &{}
      &{0.81$\pm$0.06} &{0.76$\pm$0.08} &{0.72$\pm$0.08} &{0.70$\pm$0.09} &{0.76$\pm$0.08}\\

{MPW\cite{tareef2018multi}}
      &{0.82$\pm$0.06} &{0.80$\pm$0.07} &{0.77$\pm$0.07}
      &{0.73$\pm$0.09} &{0.79$\pm$0.07} &{}
      &{0.81$\pm$0.07} &{0.79$\pm$0.09} &{0.74$\pm$0.08} &{0.71$\pm$0.10} &{0.78$\pm$0.08}\\

{CF\cite{song2018automated}}
     &{0.84$\pm$0.08} &{0.82$\pm$0.05} &{0.79$\pm$\textbf{0.06}}
     &{0.77$\pm$0.07} &{0.81$\pm$0.06} &{}
     &{0.83$\pm$0.07} &{0.81$\pm$0.08} &{0.79$\pm$0.08} &{0.75$\pm$0.09} &{0.80$\pm$0.08}\\

\midrule
{Ours-F} &{0.80$\pm$0.05} &{0.79$\pm$0.05} &{0.77$\pm$0.07}
         &{0.73$\pm$0.08} &{0.76$\pm$0.07} &{}
         &{0.80$\pm$0.06} &{0.78$\pm$0.07} &{0.76$\pm$0.09} &{0.74$\pm$0.10} &{0.77$\pm$0.07}\\

{Ours-M} &{0.81$\pm$0.07} &{0.80$\pm$0.06} &{0.79$\pm$0.07}
         &{0.74$\pm$0.09} &{0.78$\pm$0.06} &{}
         &{0.81$\pm$0.06} &{0.79$\pm$0.06} &{0.78$\pm$0.08} &{0.75$\pm$0.09} &{0.78$\pm$0.07}\\

{Ours-L} &{0.82$\pm$0.06} &{0.80$\pm$0.05} &{0.80$\pm$\textbf{0.06}}
         &{0.78$\pm$0.07} &{0.80$\pm$0.06} &{}
         &{0.81$\pm$\textbf{0.05}} &{0.80$\pm$0.07} &{0.78$\pm$0.07} &{0.77$\pm$0.09} &{0.80$\pm$0.07}\\

\midrule
{Ours} &{\textbf{0.85}$\pm$\textbf{0.04}}
       &{\textbf{0.83}$\pm$\textbf{0.04}}
       &{\textbf{0.82}$\pm$\textbf{0.06}} &{\textbf{0.81}$\pm$\textbf{0.06}} &{\textbf{0.83}$\pm$\textbf{0.05}} &{}
       &{\textbf{0.84}$\pm$\textbf{0.05}} &{\textbf{0.83}$\pm$\textbf{0.05}} &{\textbf{0.82}$\pm$\textbf{0.06}} &{\textbf{0.80}$\pm$\textbf{0.07}} &{\textbf{0.82}$\pm$\textbf{0.06}}\\
\bottomrule
\end{tabular}
\end{center}
\label{TableResultObject}
\vspace{-6pt}
\end{table*}

\subsection{Compared Methods}

We evaluate the effectiveness of the proposed method by comparing it against four state-of-the-arts: joint level set functions \cite{lu2015improved}, multiple cells labelling \cite{song2017accurate}, multi-pass watershed \cite{tareef2018multi}, and contour fragments \cite{song2018automated}, denoted respectively by LSF, MCL, MPW, and CF in the text.
Among them, LSF, MCL, and CF belong to shape prior-based methods, while MPW is a variant of watershed.
We produce the segmentation results of LSF, MCL, and CF by re-running the implementation codes provided by the authors, while the results of MPW are produced by reproducing them using the implementation provided by the authors with the recommend parameter setting.

For further evaluating the proposed method, we also conducted three ablation studies.
We denote them by Ours-F, Ours-M, and Ours-L.
Ours-F means that the built shape hypothesis set here is finite; each shape hypothesis actually is a pre-collected shape example, as suggested in \cite{song2017accurate,song2018automated}.
By conducting Ours-F, our purpose is to demonstrate the role of an infinitely large shape hypothesis set playing in the overlapping cytoplasm segmentation.
Ours-M replaces our multi-shape hypotheses selecting procedure with the level set model \cite{lu2015improved,song2017accurate}.
Its aim is to evaluate the effect of exploiting global shape priors (clump-level) and enforcing the shape constraint on the segmentation performance improvement.
Ours-L removes the shape example's importance learning procedure from our method, to demonstrate that how much extent this procedure alleviates the burden on the manual collecting of representative shape examples in the shape hypothesis set building.

\subsection{Results}

\begin{figure*}[!t]
  \centering
  \setlength{\abovecaptionskip}{0pt}
  \setlength{\belowcaptionskip}{-6pt}
  \subfigure[]{\includegraphics [width = 0.96in, height = 4.5in]{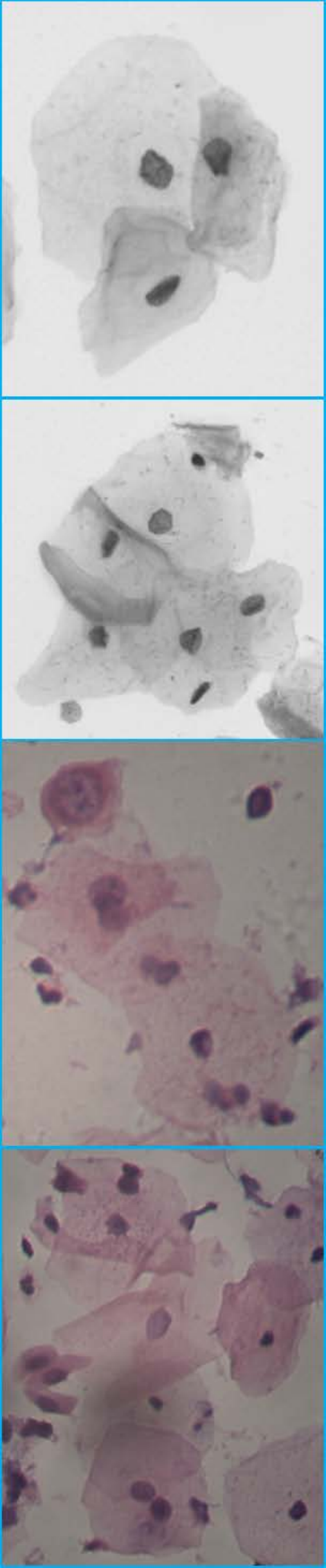}}
  \subfigure[]{\includegraphics [width = 0.96in, height = 4.5in]{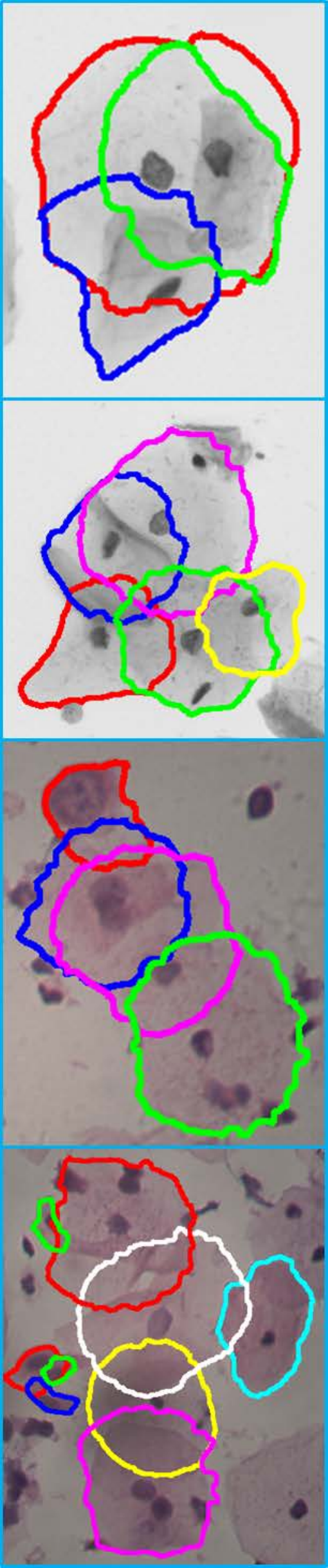}}
  \subfigure[]{\includegraphics [width = 0.96in, height = 4.5in]{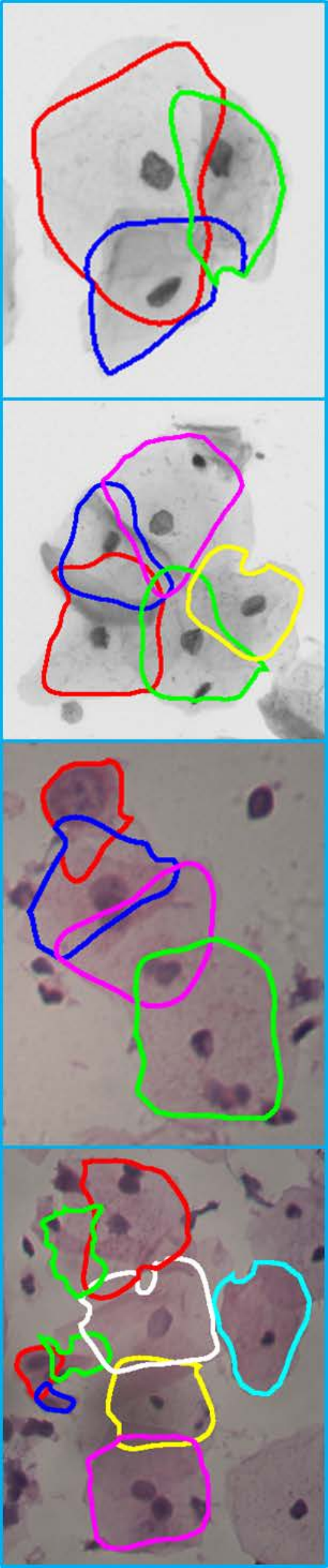}}
  \subfigure[]{\includegraphics [width = 0.96in, height = 4.5in]{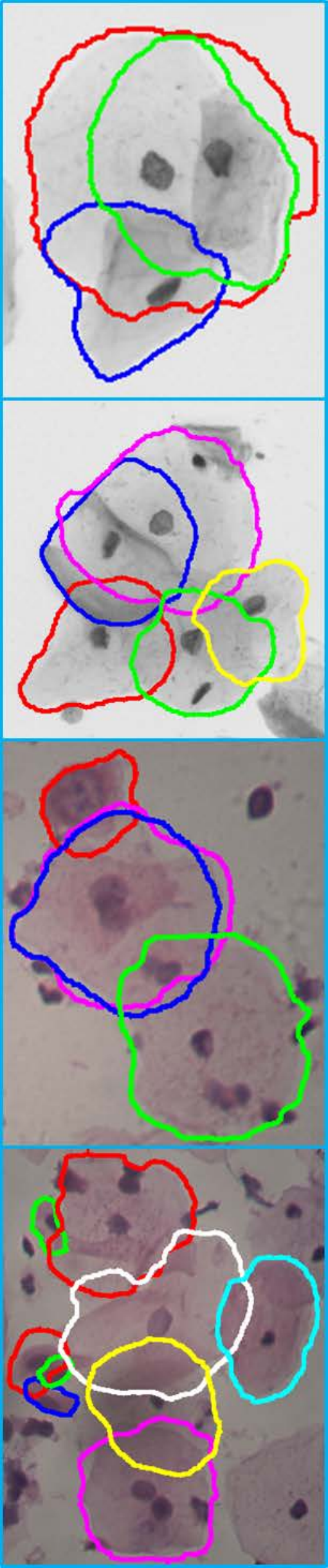}}
  \subfigure[]{\includegraphics [width = 0.96in, height = 4.5in]{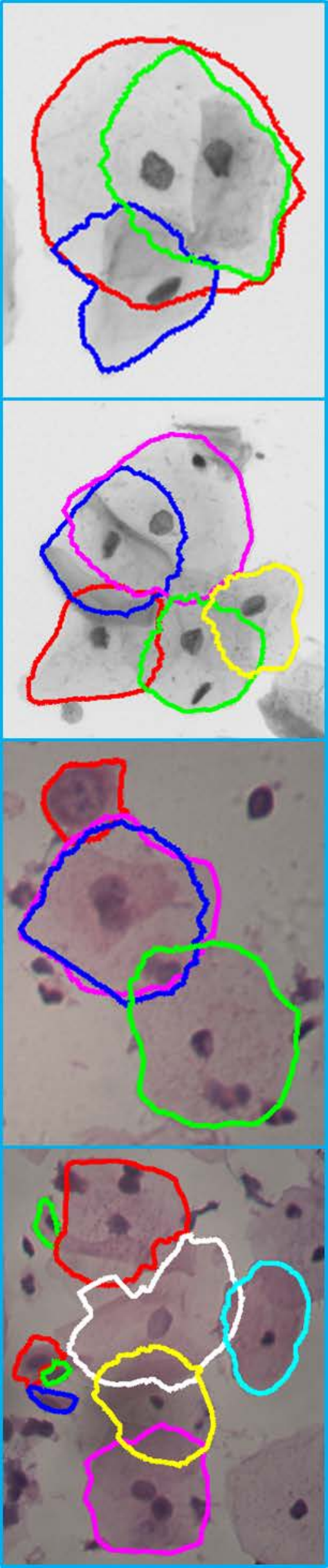}}
  \subfigure[]{\includegraphics [width = 0.96in, height = 4.5in]{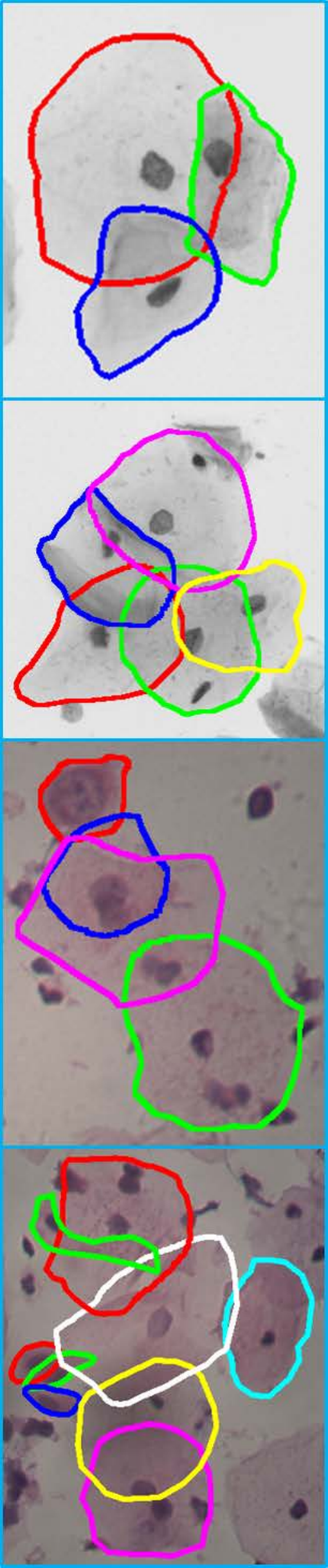}}
  \subfigure[]{\includegraphics [width = 0.96in, height = 4.5in]{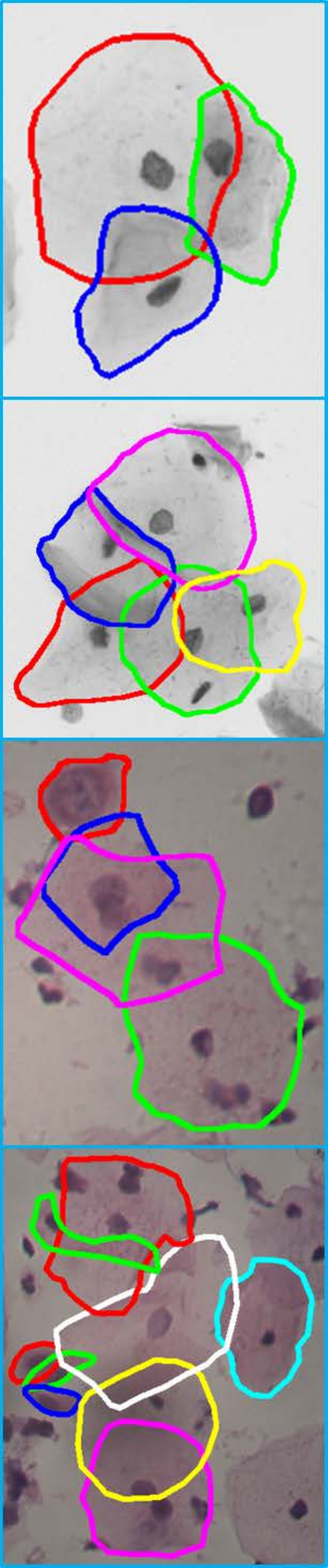}}
  \caption{Visual comparison of different segmentation methods on four examples: (a) input images, (b)--(f) results produced by LSF [Lu et al., 2015], MCL [Song et al., 2017], MPW [Tareef et al., 2018], CF [Song et al., 2018], and our method, respectively, and (g) the ground truth. Note that examples' size has been scaled for better viewing, and that the first two examples come from the Pap stain dataset while the last two examples come from the H\&E stain dataset.}
  \label{Fig.4}
\end{figure*}

\noindent\textbf{Quantitative Comparison with the State-of-the-arts:} We first check segmentation performance on pixel-level.
Results are provided in Table~\ref{TableResultPixel}.
We can see from Table~\ref{TableResultPixel} that our method produced the best segmentation result
and achieved an improvement, on average, $4.75\%$ on the Pap Stain Dataset and $5.50\%$ on the H\&E Stain Dataset against
other methods in terms of $TPR$.
We also can see that all methods have almost the same performance in terms of $TNR$ and $FPR$ while our method, MPW, and CF produce the best performance.
This is mainly due to that $TNR$ and $FPR$ in fact aim at measuring the segmentation performance on non-cytoplasm pixels.
In our application, as we mentioned before, the clump area is relatively clear, so it is relatively easy to segment these non-cytoplasm pixels.
Finally, our method also achieved the best performance in terms of $FNR$, with the improvement, on average, $5.00\%$ on the Pap Stain Dataset and $7.50\%$ on the H\&E Stain Dataset.
Note that here $TPR\neq1-FNR$, because cytoplasm pixels in the overlapping region belong to not just one cytoplasm, and there are hence counted more than once in evaluating $FNR$; it is slightly different from the conventional notion.

We now check object-level segmentation performance.
We report the results, as shown in Table~\ref{TableResultObject}, based on the `overlapping degree', because we want to evaluate how much overlapping extent the methods can deal with.
The overlapping degree is defined as the length ratio of the occluded boundary part(s) against the whole boundary of that cytoplasm.
From Table~\ref{TableResultObject}, we can see that our method produces the best segmentation result under different overlapping degrees.
More specifically, when the overlapping degree is less than $0.25$, our methods achieved an improvement of $3.00\%$ on the Pap Stain Dataset and $2.75\%$ on the H\&E Stain Dataset on average, while with the increasing of the overlapping degree, the improvement increases from $3.50\%$ to $7.75\%$ on the ap Stain Dataset and from $5.25\%$ to $8.75\%$ on the H\&E Stain Dataset, with the overall improvement of $4.25\%$ and $4.75\%$ on the two datasets respectively.

\vspace{10pt}\noindent\textbf{Visual Comparison with the State-of-the-arts:} We now look at four visual examples, as shown in Fig.~\ref{Fig.4}, for further evaluating the proposed method.
The first two examples are sampled from the Pap Stain Dataset while the last examples from the H\&E Stain Dataset.
In all four examples, the intensity deficiency issue arises; it is not easy even for human to segment the occluded boundary part(s) for at least one cytoplasm in each example.
We can see from Fig.~\ref{Fig.4} that our method achieved the best segmentation result among all the compared methods, producing a result that is comparable with the ground truth for many overlapping cytoplasms.
More specifically, the segmentation results produced by LSF, MCL, MPW, and CF are more implausible than by our method in terms of the shape.
In addition, our method can better point out those occluded boundary parts, and our results contradict the least with the clump evidence than other compared methods.

\vspace{10pt}\noindent\textbf{Ablation Study Results:} Three key ideas of our method are (1) building an infinitely large shape hypothesis set to model local shape priors, (2) selecting multi-shape hypotheses jointly to exploit global shape priors, and (3) learning shape example's importance to enhance the representation capability of the built shape hypothesis set.
We have conducted three experiments to quantitative evaluate their effect in the performance improvement; as mentioned before, we denote them by Ours-F, Ours-M, and Ours-L respectively.
The results are presented in Table~\ref{TableResultPixel} (pixel-level) and Table~\ref{TableResultObject} (object-level).
We can see from these two tables that three components are all necessary and mutually reinforcing, contributing to a performance improvement, on average, of $4.50\%$, $3.00\%$, and $4.00\%$ on $TPR$, $4.50\%$, $7.00\%$, and $2.50\%$ on $FNR$, and $6.00\%$, $4.50\%$, and $2.50\%$ on $DSC$, respectively.

\vspace{10pt}\noindent\textbf{Statistical Improvement Analysis:} We finally evaluate the statistical significance of the segmentation performance improvement of the proposed method against other methods, including the state-of-the-art methods, LSF, MCL, MPW, and CF, and the variants of the proposed method, Ours-F, Ours-M, and Ours-L.
The paired $t$-test results are presented in Table~\ref{TableStatistical}.
We can see from this table that the proposed method achieved a statistically significant improvement, all $p$-value less than $0.05$, demonstrating the effectiveness of the proposed method.

\begin{table*}[!t]\small
\setlength{\abovecaptionskip}{0pt}
\caption{Statistical Significance Results ($p$-value by Paired $t$-test) of the Performance Improvement of the Proposed Method against Other Methods.}
\vspace{-6pt}
\begin{center}
\begin{tabular}{p{0.55in} p{0.45in} p{0.45in} p{0.45in} p{0.45in}
               p{0.45in} p{0.0in} p{0.45in} p{0.45in} p{0.45in} p{0.45in} p{0.45in}}
\toprule
\ &\multicolumn{5}{c}{Pap Stain Dataset} \ &\multicolumn{5}{c}{\ \ \ \ \ \ \ \ \ \ \ \ \ \ \ \ \ H\&E Stain Dataset}\\

\cmidrule{2-6}
\cmidrule{8-12}

\ &{$TPR$}
  &{$TNR$}
  &{$FPR$}
  &{$FNR$}
  &{$DSC$} &{}
  &{$TPR$}
  &{$TNR$}
  &{$FPR$}
  &{$FNR$}
  &{$DSC$}\\

\midrule
{LSF\cite{lu2015improved}}
         &{0.002} &{0.005} &{0.001} &{0.006} &{0.009}
     &{} &{0.002} &{0.004} &{0.008} &{0.006} &{0.008}\\

{MCL\cite{song2017accurate}}
         &{0.002} &{0.006} &{0.009} &{0.007} &{0.004}
     &{} &{0.003} &{0.005} &{0.007} &{0.007} &{0.007}\\

{MPW\cite{tareef2018multi}}
         &{0.004} &{0.007} &{0.008} &{0.006} &{0.008}
     &{} &{0.003} &{0.006} &{0.006} &{0.005} &{0.006}\\

{CF\cite{song2018automated}}
         &{0.003} &{0.004} &{0.006} &{0.008} &{0.007}
     &{} &{0.004} &{0.005} &{0.005} &{0.004} &{0.006}\\

\midrule
{Ours-F} &{0.002} &{0.007} &{0.006} &{0.007} &{0.006}
     &{} &{0.006} &{0.006} &{0.004} &{0.006} &{0.004}\\

{Ours-M} &{0.001} &{0.003} &{0.008} &{0.006} &{0.004}
     &{} &{0.002} &{0.004} &{0.003} &{0.007} &{0.003}\\

{Ours-L} &{0.002} &{0.004} &{0.007} &{0.003} &{0.003}
     &{} &{0.003} &{0.005} &{0.004} &{0.006} &{0.004}\\

\bottomrule
\end{tabular}
\end{center}
\label{TableStatistical}
\vspace{-6pt}
\end{table*}

%% file: Discussion.tex
\section{Discussion}

Segmenting overlapping cytoplasms, as we mentioned before, plays a fundamental role in developing automatic screening of cervical cancer, the second-most common female-specific cancer.
Cervical cancer screening at the moment is recognized as the best way to protect female from cervical cancer, and hence our work is clinically significant because we are addressing the most difficult problem in the automatic cervical cancer screening and the proposed method addressed it significantly better than previous methods.
In addition, from an image processing view, our work is also significant, because segmenting overlapping objects is an active research topic due to the huge amount of applications in both biomedical and industrial fields while the proposed method provides a general idea and it also can be directly applied to other applications, as we will see soon.
In this section, our goal is to discuss the below issues that are closely related to the proposed method.

\begin{figure}[!t]
  \centering
  \setlength{\abovecaptionskip}{0pt}
  \setlength{\belowcaptionskip}{0pt}
  \subfigure[]{\includegraphics [width = 1.1in, height = 1.13in]{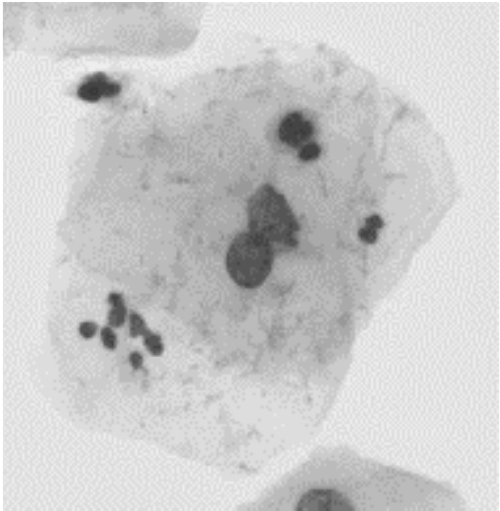}}
  \subfigure[]{\includegraphics [width = 1.1in, height = 1.13in]{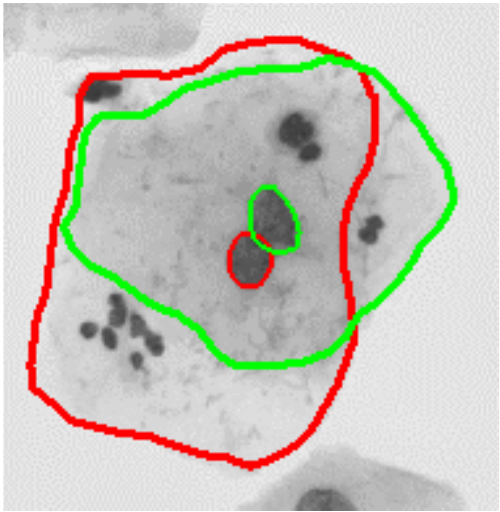}}
  \subfigure[]{\includegraphics [width = 1.1in, height = 1.13in]{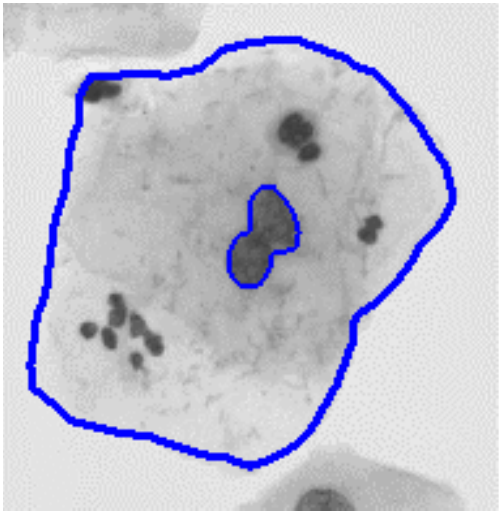}}
  \subfigure[]{\includegraphics [width = 1.1in, height = 1.13in]{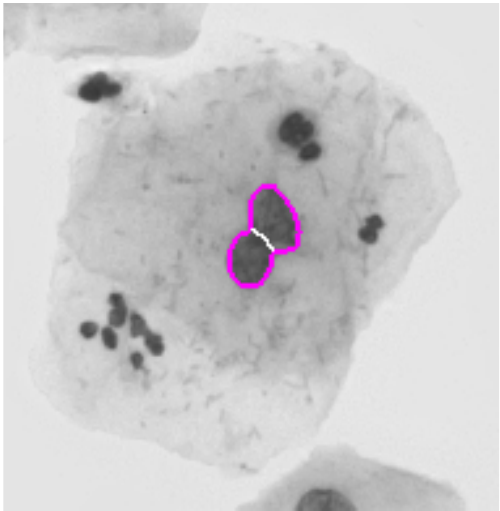}}
  \subfigure[]{\includegraphics [width = 1.1in, height = 1.13in]{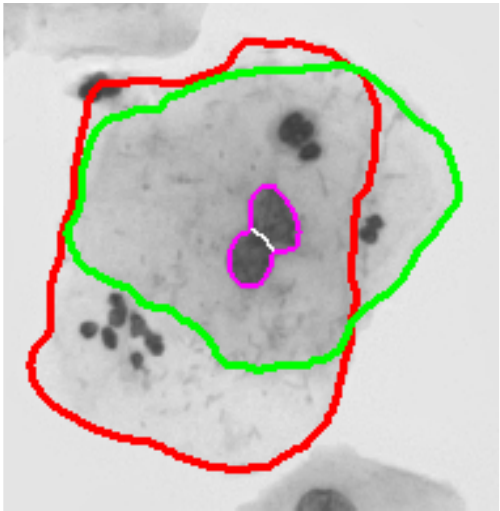}}
  \subfigure[]{\includegraphics [width = 1.1in, height = 1.13in]{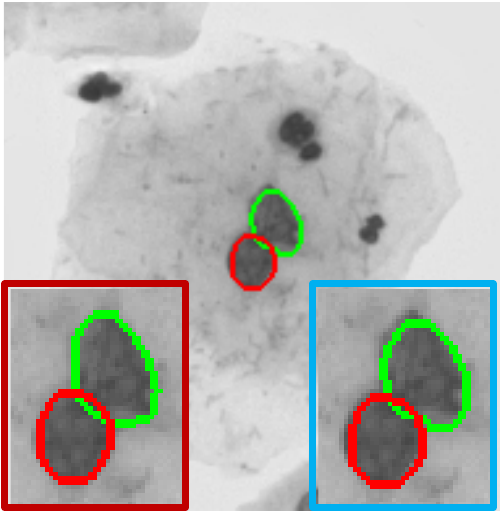}}
  \caption{A failure case: (a) input image where nuclei are overlapping, (b) the ground truth, (c) our result, (d) the splitting result of nuclei by \cite{song2015accurate} where the white line is the border, (e) our result after overlapping nuclei splitting, (f) the overlapping nuclei segmentation result by our method where for better viewing we have zoomed in the ground truth (left) and the result (right).}
  \label{Fig.5}
  \vspace{-8pt}
\end{figure}

\vspace{10pt}\noindent\textbf{Limitation:} The limitation of the proposed method derives from our assumption: nuclei in the clump are not overlapping.
This assumption holds true in the most cases, but in a few cases nuclei are indeed overlapping (see Fig.~\ref{Fig.5} for example).
When we encounter overlapping nuclei, by simply applying the proposed method, some cytoplasms will be ignored to segment, as our method treat all overlapping nuclei as a nucleus.
However, it is not a serious limitation, because we can first segment these overlapping nuclei and then employ the proposed method to segment overlapping cytoplasms.
Since cervical nucleus looks like having, more or less, a convex shape, segmenting overlapping nuclei is not so difficult as segmenting overlapping cytoplasms.
We here will not go into details of overlapping nuclei segmentation; readers who are interest in this topic can refer to \cite{song2015accurate,kong2011partition,janssens2013charisma,park2012segmentation}.
Our aim here is to point out that the limitation of the proposed method can be well addressed by employing an overlapping nuclei segmentation technique.
Fig.~\ref{Fig.5} provides such a successful example where we employ \cite{song2015accurate} to segment overlapping nuclei.
A specific treatment of how to segment overlapping nuclei to further improve the effectiveness of the proposed method is discussed below.

\vspace{10pt}\noindent\textbf{Future Work:} Although there are various works in the literature that investigate the segmentation of overlapping nuclei, they in fact are just to split the overlapping nuclei.
Their goal is to search a border that separates two overlapping nuclei, instead of segmenting the occluded boundary parts.
As a result, it is more likely leading to an inaccurate measurement on nucleus' property, which will finally degrades the accuracy of cervical cancer screening.
The proposed method originally is designed to segment occluded boundary parts.
We hence plan to generalize our method to overlapping cervical nuclei segmentation, such that we are toward more closely to the automatic cervical cancer screening.
Recall that our method just needs a centroid to segment the cytoplasm.
We hence can employ a regression technique to locate the centroid of the overlapped nucleus by regressing the distance of nuclei pixels to the centroid; such a successful technique has been reported in \cite{xie2018efficient}.
Once nucleus's centroid has been obtained, the proposed method can be directly applied to segment overlapping nuclei.
We provide such a successful example in Fig.~\ref{Fig.5} where the centroid of the nucleus currently is manually given and the built shape hypothesis set is the same as the cytoplasm.
In the future, our aim is to (1) design such a regression method to locate the centroid of the overlapped nucleus and (2) refine the proposed method to segment overlapping nuclei.

\begin{figure}[!t]
  \centering
  \setlength{\abovecaptionskip}{0pt}
  \setlength{\belowcaptionskip}{0pt}
  \subfigure[]{\includegraphics [width = 0.65in, height = 2.6in]{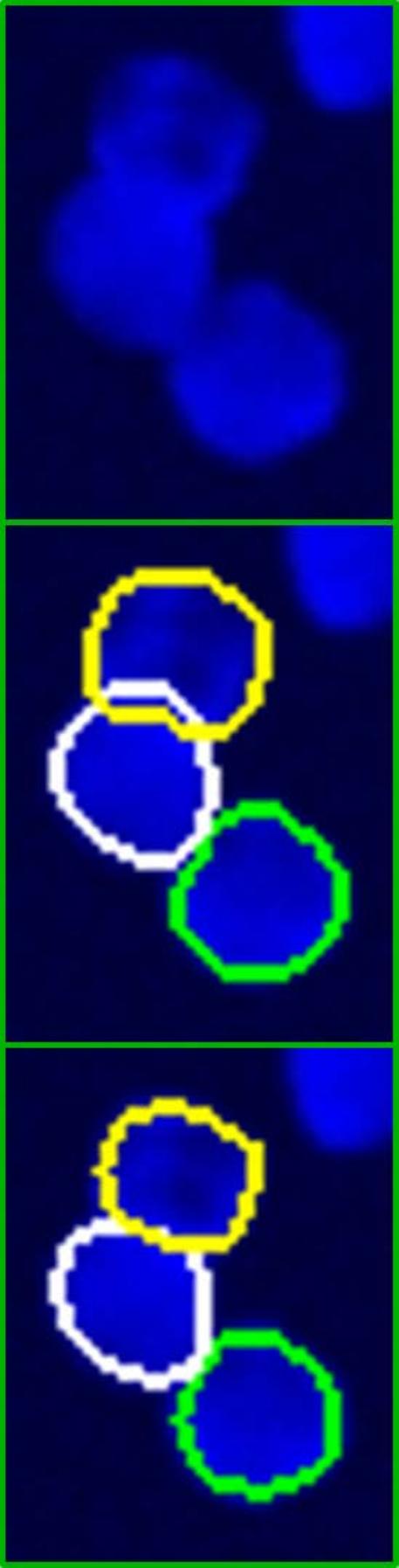}}
  \subfigure[]{\includegraphics [width = 0.90in, height = 2.6in]{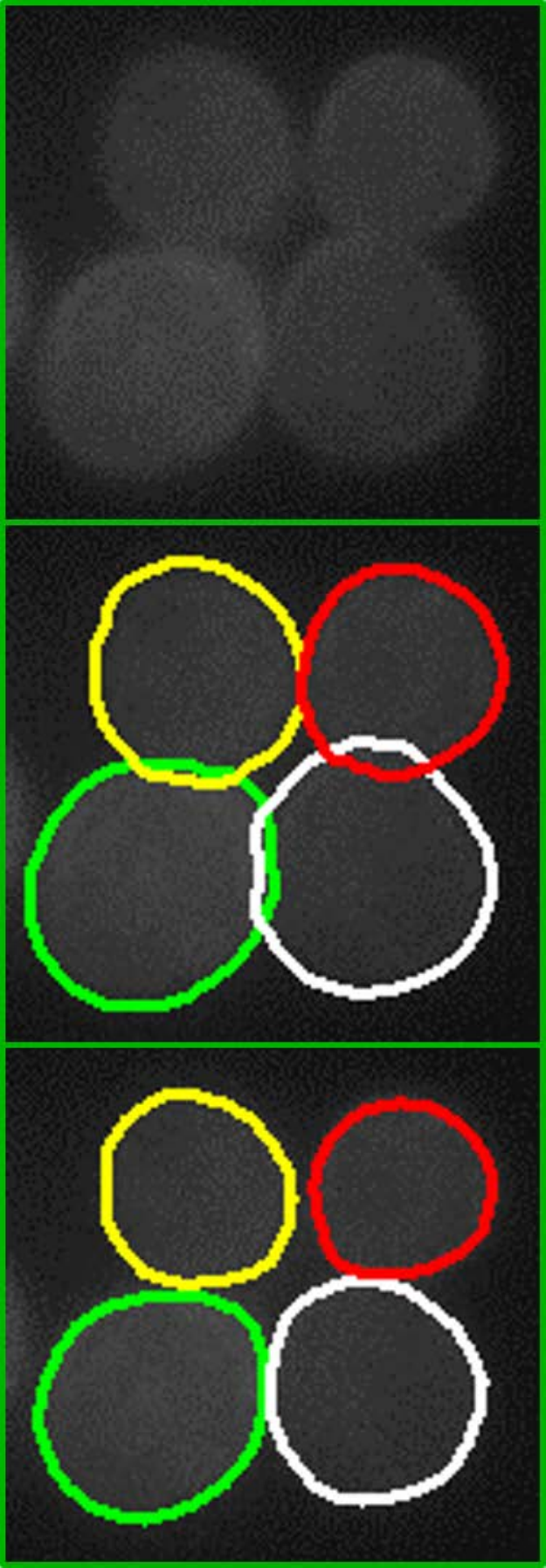}}
  \subfigure[]{\includegraphics [width = 0.94in, height = 2.6in]{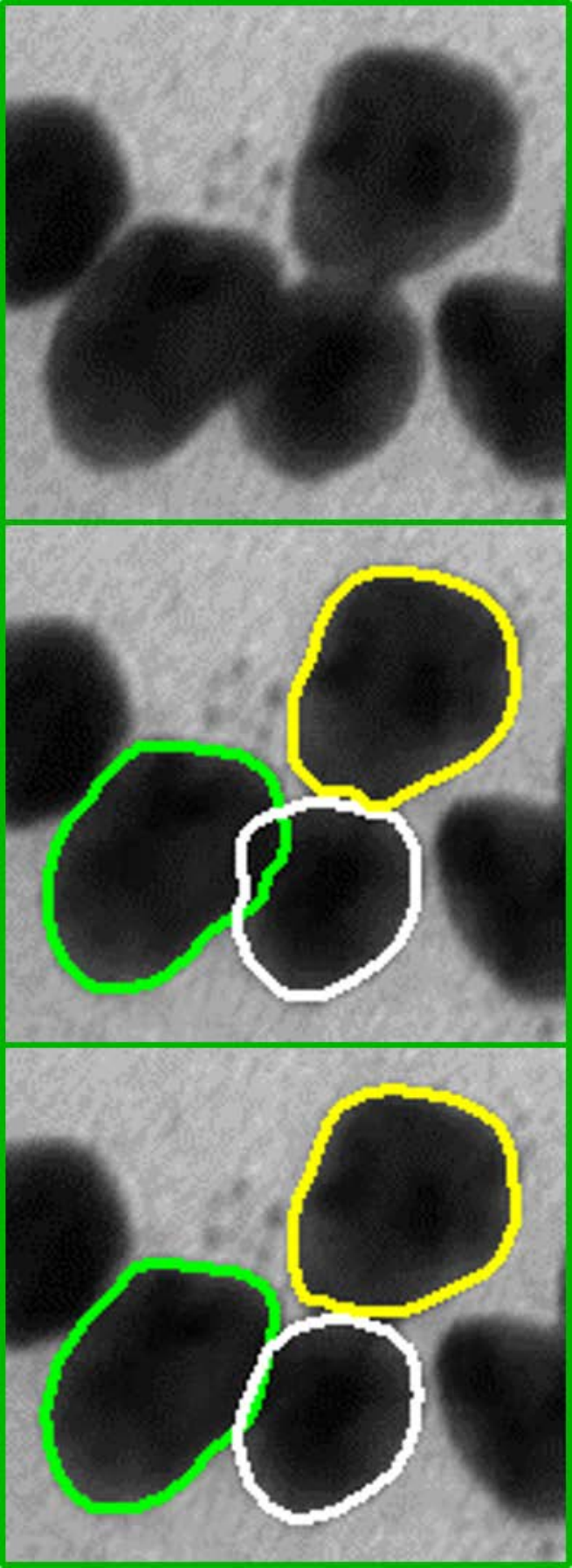}}
  \subfigure[]{\includegraphics [width = 0.81in, height = 2.6in]{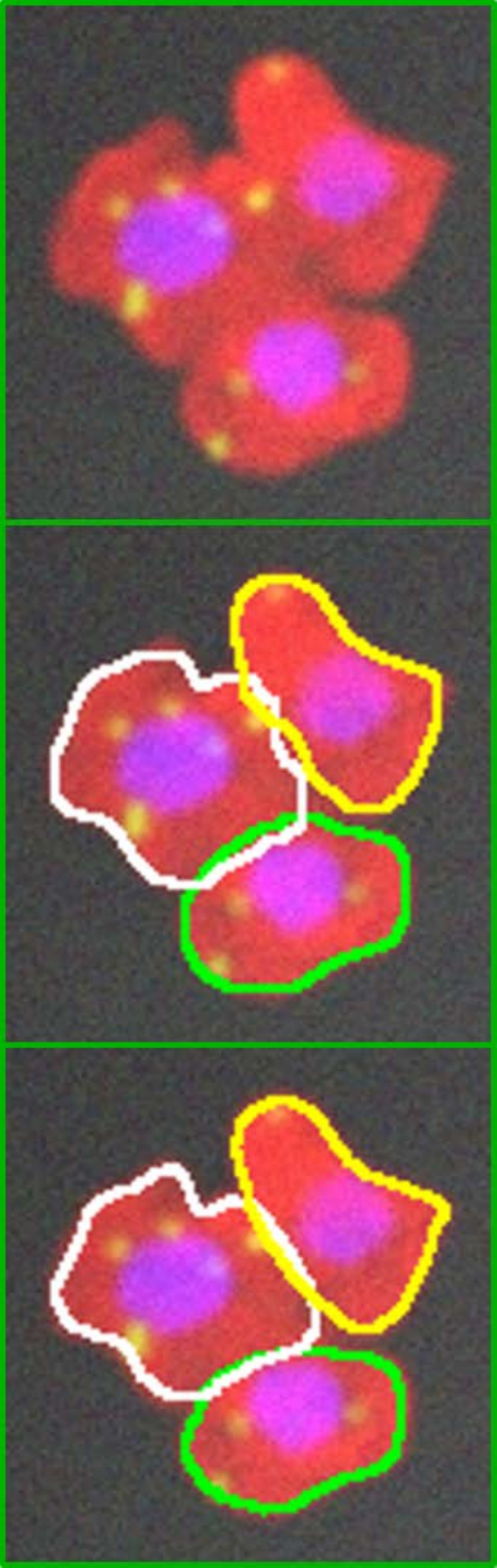}}
  \caption{Four successful examples by directly applying the proposed method to segment overlapping objects. In each sub-figure, from top to bottom row are input image, the result, and the ground truth. (a) Simulated cells (they are synthetic benchmark images provided at \protect\footnotemark[1] for validating cell image analysis algorithms), (b) Saccharomyces cerevisiae cells, (c) Nanoparticles, (d) Drosophila Kc $167$ cells.}
  \label{Fig.6}
  \vspace{-6pt}
\end{figure}

\vspace{10pt}\noindent\textbf{Applicability:} As we mentioned before, the proposed method provides a general idea of overlapping objects segmentation.
Theoretically, it can be applied to other overlapping objects segmentation tasks.
Two prerequisites of directly applying the proposed method are the clump area and the centroid of each overlapped object should be given.
Although the clump area here is well segmented by the employed multi-scale CNN \cite{song2015accurate}, it is usually depends on the application at hand; it is not impossible that we have to employ another technique to segment the clump area in other applications.
In addition, in many applications the object has not a nucleus, so we also have to locate the centroid in advance.
However, when these two prerequisites are met, we can achieve an competitive performance by directly applying our method to other overlapping objects segmentation tasks.
Fig.~\ref{Fig.6} gives four such successful examples, covering different imaging modalities and various overlapping cases.
The performance by directly comparing with its ground truth demonstrates that it is possible to directly apply or slightly refine the proposed method to other overlapping objects segmentation tasks.
Note that in all examples the clump area and the object centroid are manually given, and the shape hypothesis set is the same as the cytoplasm.

\footnotetext[1]{http://www.cs.tut.fi/sgn/csb/simcep/benchmark/}
 benchmark images presented in article Benchmark set of synthetic images for validating cell image analysis algorithms

\vspace{10pt}\noindent\textbf{Difference from the Common Pixel-wise Segmentation:} One who are not familiar with this application may wonder why we do not employ a CNN to segment overlapping cytoplasms.
After all CNN has been verified that it is able to produce a competitive performance in a wide range of segmentation tasks.
Here we have to point out that the segmentation of overlapping objects is different from the common pixel-wise segmentation tasks.
In the common pixel-wise segmentation scenario, each pixel just has one true segmentation result, while in the overlapping objects segmentation scenario pixels in the overlapping region have more than one true segmentation result, because these pixels indeed belong to different objects.
This difference substantially hinders us to employ a CNN, and is one of main reasons why we have to formulate our segmentation task as a shape hypotheses selecting problem rather than directly using the existing pixel-wise formulation.

\vspace{10pt}\noindent\textbf{Difference from Level Set Models:}
The proposed method has its similarity in spirit to level set models.
We here have to give credit to them; the nature underlying level set model (incorporating prior shape knowledge into segmentation) inspired this work.
However, there are also three significant differences that are crucial to succeed in the overlapping cytoplasm segmentation.
First, we use intensity information in a different way.
It cannot be directly used, because in the overlapping region intensity information is unreliable or even misleading to produce an accurate segmentation result.
Second, we consider all cytoplasms jointly rather than individually, which provides a better opportunity to exploit global shape priors.
Finally, we directly impose shape constraints on the resulting shape, which helps to avoid producing implausible segmentation results.
Note that in terms of implementation details the proposed method is totaly different from level set models; the formulation and optimization scheme are totally different.

%% file: Conclusion.tex
\section{Conclusion}

In this paper, we present a novel and effective method, constrained multi-shape evolution, to segment overlapping cytoplasms.
Our key idea is to exploit the clear information provided by the clump boundary while ignoring the unreliable or even misleading intensity information in the overlapping region.
To do so, we cast the segmentation problem into a shape hypotheses selecting problem, which immediately brings three advantages compared with existing methods.
We are first able to model more detailed local shape priors by building an infinitely large shape hypothesis set.
We are also able to consider global shape priors readily that now are not so difficult to formulate and incorporate into the segmentation.
Finally, it provides a better opportunity to directly constrain the resulting shape which can significantly reduce to produce the implausible segmentation results.
By conducting extensive experiments on two widely used cervical smear datasets, the proposed method is firmly verified to be able to effectively segment overlapping cytoplasms, consistently outperforming the state-of-the-art methods.

%% file: Appendix.tex
\appendices
\section{Convergence Analysis of our Learning Algorithm}

We made statements in the text that (1) our importance learning algorithm is convergent: `the importance learned by our learning algorithm is guaranteed to arrive at the right importance in the end', and (2) the convergence result holds regardless of the order of shape examples we feed: `the result holds regardless of which input-output pair $(B_c^j, \{\mathrm{\textbf{s}}_i^j\}_{i=1}^{N_j})$ we choose and which shape example $\mathrm{\textbf{s}}_i^j$ in the input-output pair we choose at each importance updating step'.
We have given an intuitive explanation why our statements hold in the text, and now we proceed to the formal proof.

\vspace{10pt}\noindent\textbf{Proof:} Recall that we denote the right importance by $\textbf{w}^*$ and the learned importance at step $t$ by $\textbf{w}(t)$.
For simplicity, we here count a step $t$ only at which the importance is updated.
Our proof, as we mentioned before, is to get a contradiction: $\cos\theta(t)>1$ if $t$ could be infinitely large.

We start by introducing three symbols: $\delta$, $\xi$, and $\mathbbm{1}_i$, where $\delta=\min\textbf{w}^*_{[i]}$, $\xi=\max\textbf{w}^*_{[i]}$, and $\mathbbm{1}_i$ has the same size with $\textbf{w}^*$ and its $i$th entry has a value of $1$ while all other entries of $0$; here $\textbf{w}^*_{[i]}$ means the value of $i$th entry of $\textbf{w}^*$.
With the definition of $\mathbbm{1}_i$, we now can summarize the difference between $\textbf{w}(t)$ and $\textbf{w}(t-1)$ as
\begin{equation}
   \Delta\textbf{w}(t)=\eta_i\ell\mathbbm{1}_i,
   \tag{A1}
\label{Eq.difference}
\end{equation}
where $\eta_i$ is the number of successful try to increase $i$th shape example's importance by $\ell$.

We now check the value of $\textbf{w}^T(t)\textbf{w}^*$.
We have
\begin{align}
   \textbf{w}^T(t)\textbf{w}^*&=\textbf{w}^T(t-1)\textbf{w}^*
   +\Delta\textbf{w}^T(t)\textbf{w}^*\notag\\&=
   \textbf{w}^T(t-1)\textbf{w}^*+\eta_i\ell\mathbbm{1}^T_i\textbf{w}^*
   \notag\\&\geq\textbf{w}^T(t-1)\textbf{w}^*+\eta_i\ell\delta.
   \tag{A2}
\label{Eq.numertator}
\end{align}
By applying above inequality $t$ times, we get
\begin{align}
   \textbf{w}^T(t)\textbf{w}^*\geq\textbf{w}^T(0)\textbf{w}^*
   +\ell\delta\sum_{i=1}^{t}\eta_i\geq t\ell\delta.
   \tag{A3}
\label{Eq.numertatorInduction}
\end{align}
On the other hand, we have
\begin{align}
   \Vert\textbf{w}(t)\Vert^2&=\Vert\textbf{w}(t-1)+
   \Delta\textbf{w}(t)\Vert^2\notag\\&=
   \Vert\textbf{w}(t-1)\Vert^2+2\eta_i\ell\textbf{w}^T(t-1)\mathbbm{1}_i
   +\eta_i^2\ell^2\Vert\mathbbm{1}_i\Vert^2\notag\\
   &\leq\Vert\textbf{w}(t-1)\Vert^2+2\eta_i\ell\xi+\eta_i^2\ell^2\notag\\
   &\leq\Vert\textbf{w}(0)\Vert^2+2\ell\xi\sum_{i=1}^{t}\eta_i+
   \ell^2\sum_{i=1}^{t}\eta_i^2\notag\\
   &\leq N+(2\ell\xi\eta+\ell^2\eta^2)t,
   \tag{A4}
\label{Eq.denominator}
\end{align}
where $N$, as we mentioned in the text, stands for the number of all cytoplasms we used during learning, and $\eta=\max\eta_i$.

Finally, by using Eq.~\ref{Eq.numertatorInduction} and Eq.~\ref{Eq.denominator}, we get
\begin{align}
   \cos\theta(t)&=\frac{\textbf{w}^T(t)\textbf{w}^*}
   {\Vert\textbf{w}(t)\Vert\Vert\textbf{w}^*\Vert}\notag\\
   &\geq\frac{t\ell\delta}{\sqrt{N+(2\ell\xi\eta+\ell^2\eta^2)t}N\xi}
   \notag\\ &\rightarrow\sqrt{t}=+\infty \text{ as } t\rightarrow{+\infty}
   \tag{A5}
\label{Eq.contradiction}
\end{align}
and complete the proof.
\vspace{-12pt}\begin{flushright}\tiny
$\square$\
\end{flushright}

We below derive the convergence bound.
First note that $N$ is a constant when the training dataset has been collected, and that $t$ is finite.
Hence $N$ in Eq.~\ref{Eq.denominator} can be replaced with $\alpha t$ where $\alpha$ denotes the unknown ratio between $N$ against $t$.
We then have
\begin{align}
   \frac{(\textbf{w}^T(t)\textbf{w}^*)^2}{\Vert\textbf{w}(t)\Vert^2}
   &\geq\frac{(t\ell\delta)^2}{\alpha t+(2\ell\xi\eta+\ell^2\eta^2)t},
   \tag{A6}
\label{Eq.bound1}
\end{align}
which implies
\begin{align}
   t&\leq\frac{\alpha+2\ell\xi\eta+\ell^2\eta^2}{(\ell\delta)^2}
   \frac{(\textbf{w}^T(t)\textbf{w}^*)^2}{\Vert\textbf{w}(t)\Vert^2}
   \notag\\&=\frac{\alpha+2\ell\xi\eta+\ell^2\eta^2}{(\ell\delta)^2}
   \frac{(\textbf{w}^T(t)\textbf{w}^*)^2}{\Vert\textbf{w}(t)\Vert^2}
   \frac{\Vert\textbf{w}^*\Vert^2}{\Vert\textbf{w}^*\Vert^2}\notag\\
   &\leq\frac{\alpha+2\ell\xi\eta+\ell^2\eta^2}{(\ell\delta)^2}
   \Vert\textbf{w}^*\Vert^2\notag\\
   &\leq N\frac{\alpha+2\ell\xi\eta+\ell^2\eta^2}{(\ell\delta)^2}\xi^2.
   \tag{A7}
\label{Eq.bound2}
\end{align}
Here note that by Cauchy-Schwarz inequality \cite{bhatia1995cauchy} we have
\begin{align}
   (\textbf{w}^T(t)\textbf{w}^*)^2\leq\Vert\textbf{w}(t)
   \Vert^2\Vert\textbf{w}^*\Vert^2,
   \tag{A8}
\label{Eq.bound3}
\end{align}
which gives the result in Eq.~\ref{Eq.bound2}.